\documentclass[11pt]{article}
\usepackage[margin=0.91in]{geometry} 
\usepackage{kpfonts}
\usepackage[colorlinks,
citecolor=blue
]{hyperref}  
\usepackage{url}
\usepackage{bm}
\usepackage{microtype}
\usepackage{graphicx}
\usepackage{subfigure}  
\usepackage{booktabs} 

\usepackage{hyperref}
\usepackage{algorithm, algorithmic}

\usepackage{natbib}

\usepackage{multirow}
\usepackage{colortbl}
\usepackage{booktabs,multirow,array,caption}
\usepackage{pbox}
\usepackage{pifont}
\usepackage{tikz}
\usepackage{xcolor}
\usepackage{graphicx}

\DeclareSymbolFont{extraup}{U}{zavm}{m}{n}
\DeclareMathSymbol{\varheartsuit}{\mathalpha}{extraup}{86}
\DeclareMathSymbol{\vardiamondsuit}{\mathalpha}{extraup}{87} 

\usepackage{amsfonts}
\usepackage{xcolor}
\let\proof\relax
\let\endproof\relax
\usepackage{amsthm}
\usepackage{amsmath}
\usepackage{amssymb}
\usepackage{soul}
\usepackage{comment}
\usepackage{enumitem}
\usepackage{bbm}
\usepackage{graphicx}
\usepackage{float}
\usepackage{mathtools}

\newtheorem{definition}{Definition}

\theoremstyle{remark}

\renewcommand{\epsilon}{\varepsilon}

\usepackage{array}
\newcolumntype{M}[1]{>{\centering\arraybackslash}m{#1}}

\usepackage{tablefootnote}
\usepackage[symbol]{footmisc}

\title{Interactive Medical Image Segmentation with Self-Adaptive \\ Confidence Calibration}

\makeatletter
\def\@fnsymbol#1{\ensuremath{\ifcase#1\or  \natural \or \dagger\or * \or \ddagger\or
		\mathsection\or \mathparagraph\or \|\or **\or \dagger\dagger
		\or \ddagger\ddagger \else\@ctrerr\fi}}
\makeatother

\author{\normalsize Wenhao Li\thanks{School of Computer Science and Technology, East China Normal University, Shanghai, China. Email addresses: \{52194501026@stu, 51184501067@stu, 52205901007@stu, bjin@cs, xfwang@cs\}.ecnu.edu.cn. The first four authors contribute equally.} \and Qisen Xu$^\natural$ \and Chuyun Shen$^\natural$ \and  Bin Hu\thanks{Huashan Hospital, Fudan University, Shanghai, China. Email: \{08301010188, zhufengping, liyuxin\}@fudan.edu.cn} \and Fengping Zhu$^\dag$ \and Yuxin Li$^\dag$ \and Bo Jin$^\natural$
\and Xiangfeng Wang$^\natural$
}

\date{\normalsize }

\begin{document}
 
 \maketitle

\begin{abstract}
Medical image segmentation is one of the fundamental problems for artificial intelligence-based clinical decision systems. 
Current automatic medical image segmentation methods are often failed to meet clinical requirements.
As such, a series of interactive segmentation algorithms are proposed to utilize expert correction information.
However, existing methods suffer from some segmentation refining failure problems after long-term interactions and some cost problems from expert annotation, which hinder clinical applications.
This paper proposes an interactive segmentation framework, called interactive {\bf{ME}}dical segmentation with self-adaptive {\bf{C}}onfidence {\bf{CA}}libration ({\bf{MECCA}}), by introducing the corrective action evaluation, which combines the action-based confidence learning and multi-agent reinforcement learning (MARL).
The evaluation is established through a novel action-based confidence network, and the corrective actions are obtained from MARL. 
Based on the confidential information, 
a self-adaptive reward function is designed to provide more detailed feedback, and a simulated label generation mechanism is proposed on unsupervised data to reduce over-reliance on labeled data.
Experimental results on various medical image datasets have shown the significant performance of the proposed algorithm.
\end{abstract}

\section{Introduction}
\label{introduction}
\begin{figure}[ht!]
\centering
\includegraphics[width=0.8\textwidth]{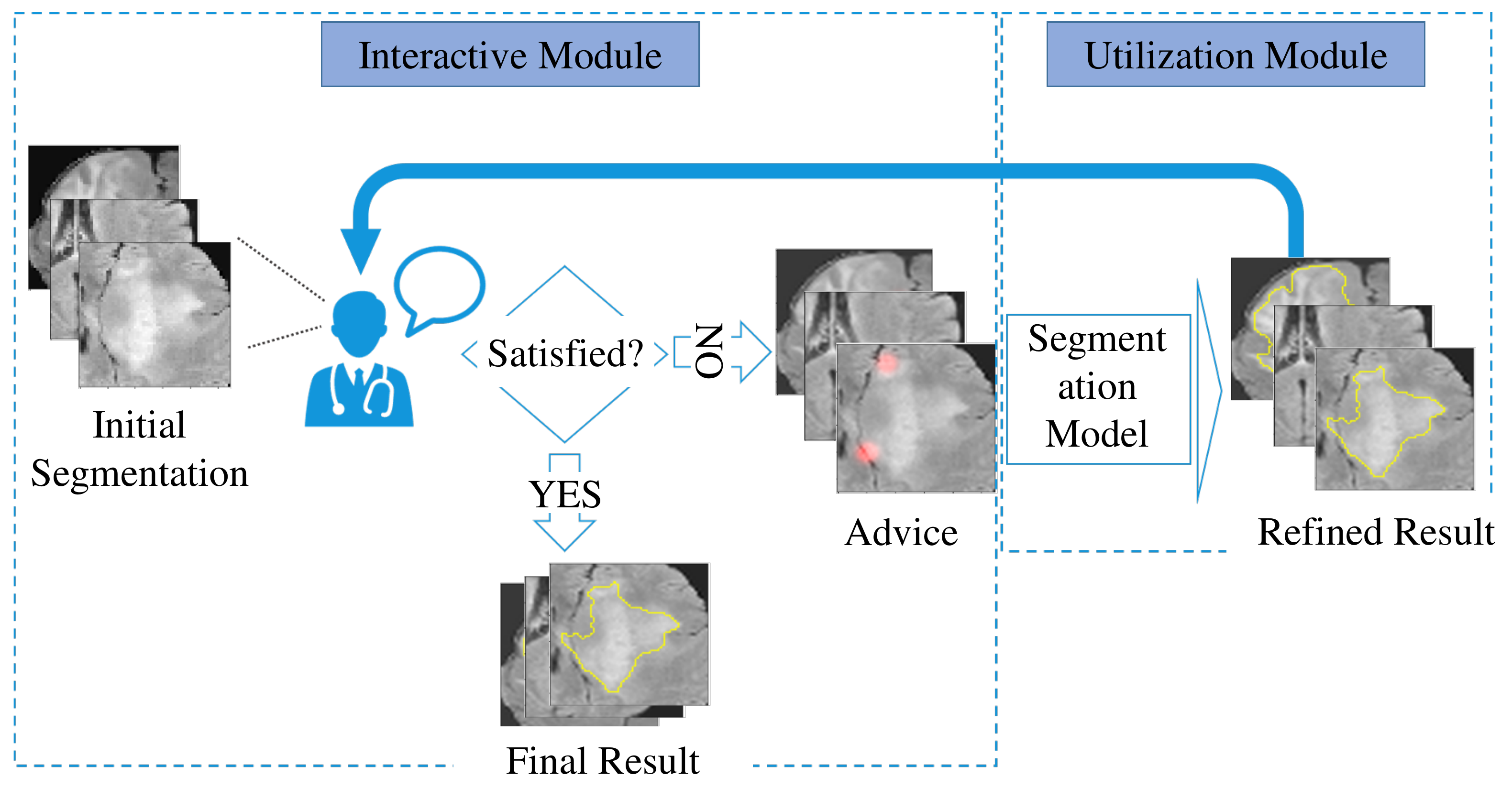}
\caption{The framework of the interactive segmentation process. Interactive module: the expert observes current (or initial) segmentation and provides further correction information (red hints); Utilization module: new segmentation is refined based on the correction information.}
\label{interaction}
\end{figure}

Medical image segmentation is one of the essential tasks for computer-aided medical diagnosis.
However, due to the pathological variability, dark lesion areas, as well as the uneven quality of the training data (lacking the consistency between imaging scanners, operators, and annotators), the accuracy of traditional convolutional neural networks (CNNs) type segmentation algorithms usually fail to meet clinical demands~\cite{wang2018interactive,wang2018deepigeos,liao2020iteratively}.
To further refine the relatively inaccurate segmentation results, interactive image segmentation algorithms that take advantage of interactive correction information ({\em{e.g.}}, clicks, scribbles, or bounding boxes) are introduced~\cite{crajchl2016deepcut,xu2016deep,lin2016scribblesup,wang2018deepigeos,wang2018interactive,bredell2018iterative,liao2020iteratively,ma2020boundary-aware}.
The general interactive segmentation process is depicted in Figure~\ref{interaction}, which contains two modules, {\em{i.e.}}, interactive module and utilization module.
In the interactive module, the users (or experts) provide some interaction correction information ({\em{e.g.}}, hint information like clicks and scribbles).
In the utilization module, the segmentation algorithm takes advantage of the interaction correction information to refine the previous model efficiently.
The interactive segmentation algorithms can achieve better performance than traditional segmentation algorithms by utilizing the additional interactive correction information.
The interaction process can be considered as the continuous cooperation between the model and the human expert.
As a result, an excellent interactive segmentation model should understand experts' interactive information and update itself to collaborate better.

\begin{figure}[htbp]
\centering
\subfigure[Correction information is ignored] { \label{badcase_b} 
\includegraphics[width=0.45\textwidth]{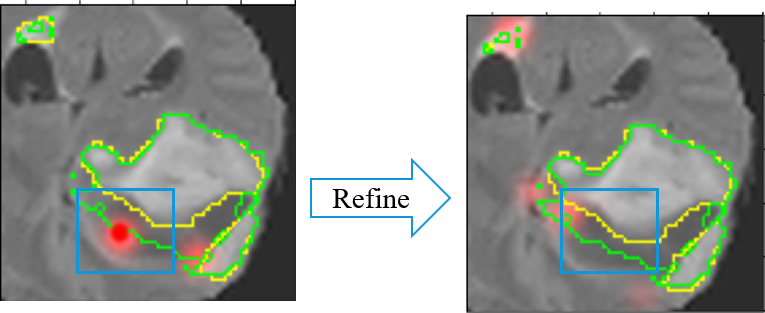}
}
\subfigure[Refined result becomes worse] { \label{badcase_a} 
\includegraphics[width=0.45\textwidth ]{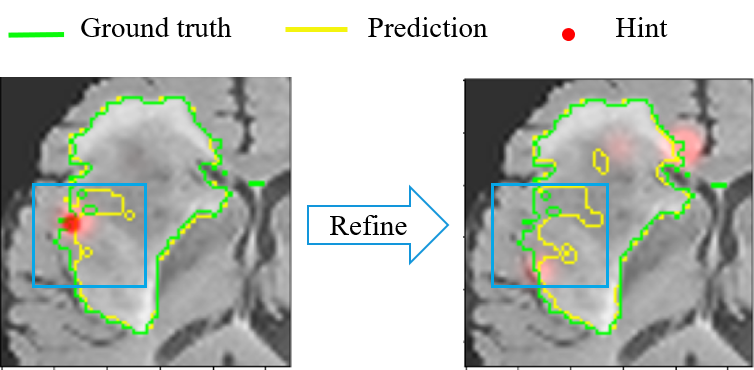} 
}
\caption{Segmentation refining failure after long-term interactions on BraTS2015 images:
a) The segmentation model can not fully understand the hint information or just ignores it.
b) The interactive segmentation model could misunderstand the expert's interaction correction, which results in the worse result;}
\label{badcase}
\end{figure}

Methods such as BIFSeg~\cite{wang2018interactive} and DeepIGeos~\cite{wang2018deepigeos} model user interactions as hard constraints through conditional random fields (CRFs) end-to-end training.
However, these model more focus on the \textit{one-step} interaction and the refined segmentation after the first step cannot be efficiently used in these model~\cite{liao2020iteratively,ma2020boundary-aware}.
Thus, these model can not utilize the long-term interactive information but the short-term interactive information efficiently.
The follow-up InterCNN~\cite{bredell2018iterative} and other methods~\cite{liao2020iteratively,ma2020boundary-aware} are improved on DeepIGeos and BIFSeg, modeling the problem as an \textit{iterative} interaction problem, which are more focus on the \textit{multi-step} interaction.
Therefore, these method can cover the data distribution corresponding to different interaction levels in the training phase to effectively use long-term interactions, while ignoring the stochasticity or uncertainty of the model makes it difficult for them to effectively use short-term interactive information.
Existing interactive methods cannot effectively utilize both short-term and long-term interactive information at the same time, which leads to the \textit{interactive misunderstanding} phenomenon\footnote{This phenomenon will be discussed in detail in the Section~\ref{action confidence}.}.
Figure \ref{badcase} presents this phenomenon
when implementing the popular interactive segmentation algorithm InterCNN~\cite{bredell2018iterative} on the BraTS2015 dataset.
The algorithm ignores the expert's correction information as in Figure \ref{badcase_b}, and even be adversely affected by correction information as in Figure \ref{badcase_a}.
These inconsistencies between hint information and the refined results indicate that existing 
algorithms still face the critical challenge of inefficiency utilization of interactive correction information.
The reason for segmentation refining failure is that, at the end of the interactive procedure, the total loss of the model will guarantee the main area's priority and ignore some small but challenging areas, such as edges.
The regions that are insensitive to interactive correction information could be considered hard-to-segment regions~\cite{nie2019difficulty,7780458}, which leads to mediocre segmentation refinement.
If these regions are not paid more attention during the training stage, the segmentation model will inevitably dominate those easy-to-segment regions.
This problem is more serious for medical images, where the hard-to-segment regions usually are tumor boundaries and are very important for clinical diagnosis and surgery.
Therefore, it becomes urgent to improve the utilization of correction information, especially for the hard-to-segment regions.


Besides, the requirement of large amounts of expert-annotated images is another crucial issue.
On the one hand, accurate annotations of medical images require plenty of time of experts, so it is expensive to acquire sufficient data with high-quality annotations. 
On the other hand, more unlabeled images can be obtained with much less cost but existing
popular interactive segmentation methods usually 
ignore utilizing these low-cost images.
Thus it becomes more and more critical to reduce the need for expert-annotated images while simultaneously taking advantage of those unlabeled images.

\begin{figure*}[ht!]
\centering
\subfigure[The architecture of MECCA.]{
\label{fig:framework}
\includegraphics[width=0.95\textwidth]{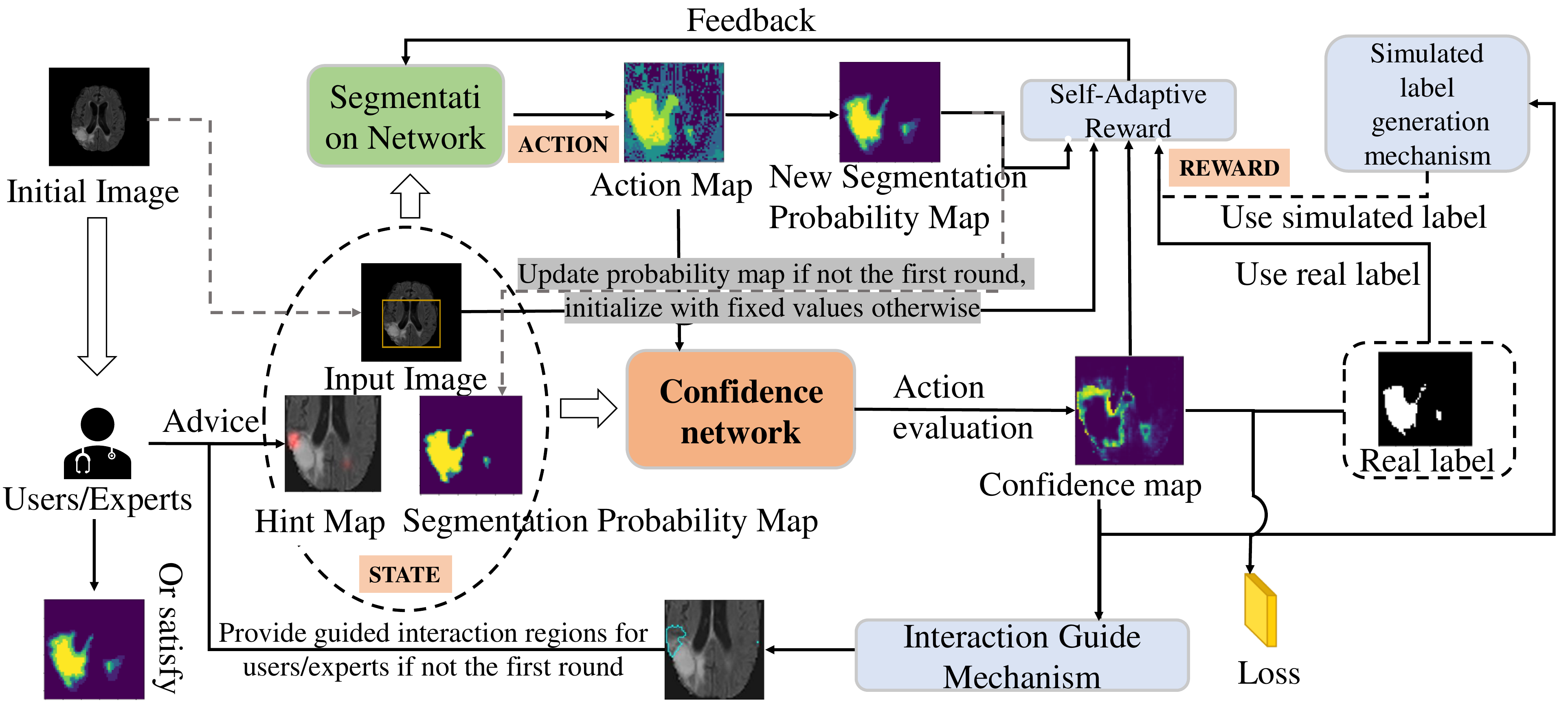}}
\hspace{1cm}
\subfigure[The testing stage of MECCA.]{
\label{testing}
\includegraphics[width=0.85\textwidth]{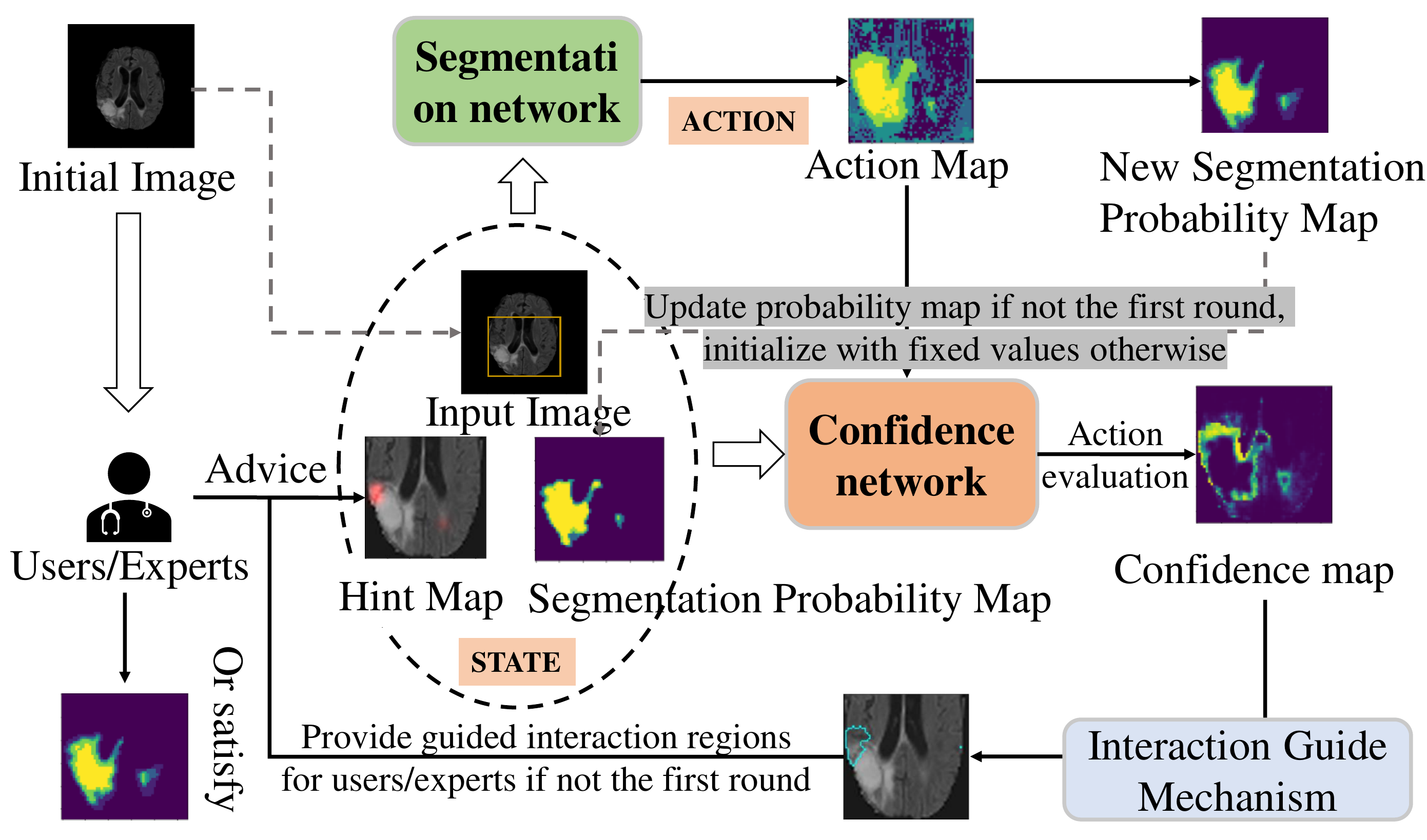}}
\caption{\textbf{(a)} The architecture of MECCA. The segmentation module outputs actions to change the segmentation probability of each voxel(agent) at each interaction step. Meanwhile, the confidence network will estimate the confidence of actions, which will generate the self-adaptive reward and simulated label. The confidence map can provide the advice regions of the next interaction step to experts. \textbf{(b)} The testing stage of MECCA. In the first instance, all voxels in the segmentation probability map are initialized to a fixed value (set to $0.5$ in the experiment). Users or experts randomly mark hint points according to the initialized segmentation probability map.}
\end{figure*}

In this paper, we propose a novel interactive segmentation algorithm for 3D medical images called interactive {\bf{ME}}dical segmentation with self-adaptive {\bf{C}}onfidence {\bf{CA}}libration (MECCA). 
MECCA combines an action-based confidence learning network with the multi-agent reinforcement learning framework.
The action-based confidence network could evaluate the corrective actions' quality by directly calculating the refinement actions' confidence.
Unlike learning the confidence of the overall segmentation result, the confidence of actions could better evaluate whether the segmentation model has correctly utilized the experts' interactive correction information.
We formulate the iterative interactions process as a Markov decision process(MDP) to model the dynamic process and further introduce the reinforcement learning technique~\cite{song2018seednet}.
Further, instead of setting each image or each patient (a series of images) as the agent, we consider each voxel\footnote{The smallest unit in three-dimensional space.} as an agent while each agent aims to learn the segmentation policy and makes its decision~\cite{furuta2019pixelrl}.
Specifically, after receiving the interactive correction information, each agent will modify its label by changing (increasing or decreasing) the category probability.
The novel action-based confidence network will directly evaluate each agent's corrective action obtained from the corrective policy of MARL.
By utilizing this action-based confidence network, two additional techniques are further proposed to improve the utilization efficiency of the interactive correction information: 
1) compared with former manual-designed rewards, a self-adaptive reward function of each action is constructed, which could provide more meticulous feedback under a flexible framework; 
2) a simulated label generation mechanism is established by utilizing the interactive correction information as the weakly supervisory signal. 
By combining the confidence network, the simulated label generation mechanism can approximately generate the labels for unsupervised images and reduce over-reliance on labeled data.
The overall algorithm framework of our proposed algorithm is shown in Figure~\ref{fig:framework}.
The main contributions of this work are summarized as follows: 1) A novel framework is proposed for interactive medical image segmentation by combining action-based confidence learning network with multi-agent reinforcement learning;
2) A self-adaptive feedback mechanism (self-adaptive reward) is constructed with the action-based confidence network to alleviate the effect of interactive misunderstanding phenomenon during the interaction process;
3) Simulated supervisory signals can be generated based on the confidence learning network and actions; hence much less labeled data (ground truth data) are needed to achieve the same performance.

The following is the roadmap of this paper.
Section \ref{related work} provides a brief but complete introduction to related works in image segmentation and interactive image segmentation, while the background material on confidence learning is also provided. 
Section \ref{method} describes the proposed interactive image segmentation algorithm, including the segmentation policy, the evaluation of corrective actions, the self-adaptive rewards, and the label generation scheme.
Extensive experiments are presented in Section \ref{exp}.
Conclusions and future work discussions are proposed in Section \ref{conclusion}.

\section{Related Works}\label{related work}
Image segmentation is the fundamental problem of computer vision or image processing that has been widely and long-termly studied.
Deep learning (DL) has further promoted the development of automatic segmentation algorithms, the same as other applications.
CNN-type methods are typical DL algorithms for image segmentation, e.g., fully convolutional networks (FCN) ~\cite{long2015fully} and DeepLab ~\cite{chen2017deeplab}.
The U-Net~\cite{ronneberger2015u}, which is considered an evilutionary variant of FCN, becomes one of the SOTA methods and performs better for medical image segmentation.
Medical image segmentation is the key to modern auxiliary diagnosis and treatment response evaluation.
A series of related works were proposed with the progressive performance for medical image segmentation ~\cite{milletari2016v,kamnitsas2017efficient,li2017compactness}.
In the following, we will review the development of interactive image segmentation methods and discuss confidence learning for image segmentation relevant to our proposed algorithm.

\paragraph{Traditional interactive image segmentation methods.}
The classical random Walk~\cite{grady2006random} can create a weight map with pixels as vertices and segment the image based on user interactions.
GrabCut~\cite{rother2004grabcut} and GraphCut~\cite{boykov2001interactive} could associate image segmentation with the maximum flow and minimum cut algorithms on graphs, respectively, while Geos ~\cite{criminisi2008geos} was proposed to measure the similarity between pixels geodesic distance.
These traditional interactive image segmentation methods aim to utilize additional expert interaction information to modify the segmentation performance further.


\paragraph{DL-based interactive image segmentation methods.}
\cite{xu2016deep} segments images based on CNN interactively.
DeepCut~\cite{crajchl2016deepcut} and ScribbleSup~\cite{lin2016scribblesup} both employed weakly supervised expert hints to establish interactive image segmentation methods.
DeepIGeoS~\cite{wang2018deepigeos} employed geodesic distance metric to construct a hint map.
The interactive segmentation process can be considered as a sequential iterative process.
It becomes natural to introduce the RL framework to model the interactive segmentation process.
Polygon-RNN~\cite{castrejon2017annotating} fundamentally segmented each target as a polygon and iteratively chose the polygon vertexes through a recurrent neural network(RNN).
Polygon-RNN+~\cite{acuna2018efficient} employed almost the same idea of Polygon-RNN but learned to choose vertexes by RL.
SeedNet~\cite{song2018seednet} trained an expert interaction generation RL model, which obtains new simulated interaction information at each interaction step.
IteR-MRL~\cite{liao2020iteratively} and BS-IRIS~\cite{ma2020boundary-aware} both modeled the dynamic interaction process as an MDP and employed multi-agent RL(MARL) models to segment images. Some researches also aim to reduce the annotation cost of interactive image segmentation. IFSL \cite{9358206} introduces interactive learning into the few-shot learning strategy and addresses the annotation burden of medical image segmentation models. IOG~\cite{9157733} proposes a practical Inside-Outside Guidance approach for minimizing the labeling cost.
These interactive methods are difficult to effectively utilize experts' short-term and long-term interaction information simultaneously, thus making error correction operations.

\paragraph{Uncertainty estimation for image segmentation}
Uncertainty estimates are helpful in the context of deployed machine learning systems as they are capable of detecting when a neural network is likely to make an incorrect prediction or when the input may be out-of-distribution.
Traditionally, much of the work are inspired by Bayesian statistics, or Bayesian Neural Network (BNN)~\cite{mackay1992practical,neal2012bayesian}.
Unfortunately, Bayesian inference is computationally intractable in practice, so much effort has been put into developing approximations of Bayesian neural networks that are easier to train.
Recent efforts to approximate the BNNs in this area include Monte-Carlo Dropout~\cite{gal2015dropout}, Multiplicative Normalizing Flows~\cite{louizos2017multiplicative}, and Stochastic Batch Normalization~\cite{atanov2019uncertainty}.
These methods are capable of producing uncertainty estimates, although with varying degrees of success.
The main disadvantage with these BNN approximations is that they require sampling to generate the output distributions.
As such, uncertainty estimates are often time-consuming or resource-intensive to produce, requiring $10$ to $100$ forward passes through a neural network to produce useful uncertainty estimates at inference time.
An alternative to BNNs is ensembling methods~\cite{dietterich2000ensemble,kamnitsas2017ensembles,mehrtash2018automatic,lakshminarayanan2016simple,mehrtash2020confidence}, which propose a frequentist approach to the problem of uncertainty estimation by training many models and observing the variance in their predictions.
However, this technique is still quite resourcing intensive, as it requires inference from multiple models to produce the uncertainty estimate.
A promising alternative to sampling-based methods is to instead have the neural network learn what its uncertainty should be for any give input, i.e., \textit{learning-based uncertainty estimation} or \textit{confidence learning}, as demonstrated in~\cite{kendall2017uncertainties,devries2018learning,robinson2018real,devries2018leveraging,jungo2019assessing,moeskops2017adversarial,hung2018adversarial,nie2019difficulty}.
These methods commonly consist of a segmentation and confidence network and are more computationally efficient than other techniques.
Thus they are better suited when computational resources are limited or when real-time inference is required, such as the interactive segmentation scenario considered in this paper.
Specifically, \cite{kendall2017uncertainties} introduces a confidence network to predict the \textit{aleatoric} and \textit{epistemic} uncertainties by imitating classic Bayesian tools.
The segmentation network of \cite{devries2018learning,devries2018leveraging} produces two separate outputs: prediction probabilities and a confidence estimate.
Confidence estimates are motivated by interpolating between the predicted probability distribution and the target distribution during training, where the degree of interpolation is proportional to the confidence estimate. 
A series of work~\cite{nie2019difficulty,hung2018adversarial,moeskops2017adversarial} focus on incorporating the uncertainty estimation into the adversarial learning process, where the segmentation network corresponds to the generator, and the confidence network is the discriminator accordingly.
\cite{moeskops2017adversarial} firstly employed GANs to improve the CNN-based brain MRI segmentation method.
The semi-supervised learning technique is used in~\cite{hung2018adversarial} to predict trustworthy regions in unlabeled images.
\cite{nie2019difficulty} proposed a difficulty-aware attention mechanism to handle those hard samples or challenging regions.
Different from the previous work to learning uncertainty through imitation, joint training, or adversarial learning, a simple but powerful alternative is to introduce an auxiliary task, such as to predict the overlap between a proposed segmentation and its ground truth~\cite{robinson2018real}, or to predict the voxel-wise false positive and false negative~\cite{jungo2019assessing}.\\

\noindent{\bf{Remark.}} In our proposed algorithm, the confidence network should evaluate the confidence of calibrating actions instead of the segmentation result, which is the most significant difference between our interactive method and previous learning-based uncertainty estimation methods.
Therefore, we design a novel action-oriented auxiliary task to predict whether the \textit{direction} of voxel-wise action is consistent with the ground truth.

\section{The Proposed Algorithm}\label{method}
This section will introduce our proposed interactive segmentation algorithm MECCA, which can iteratively evaluate the refinement actions and feedback to the segmentation model.
The algorithm framework follows the multi-agent reinforcement learning structure (Section~\ref{MARL} and Section~\ref{seg train}) and an action-based confidence learning module (Section~\ref{action confidence}) is introduced to evaluate the confidence of corrective actions.
This action-based confidence learning module can be used to establish the self-adaptive reward scheme and simulated label generation mechanism to utilize the interactive correction information efficiently.
The architecture overview of MECCA has been depicted in Figure \ref{fig:framework}.
The model's state information includes the original 3D image, the previous segmentation probability, and the hint map generated from interaction and confidence maps.
Based on the current state information, the segmentation module gives suggested actions to refine previous segmentation results by adjusting the segmentation probability of each voxel (agent).
Further, the state information will be utilized to evaluate the confidence of obtained actions through the confidence network, with a confidence map as the output.
The self-adaptive reward is designed through a self-adaptive weighting scheme based on the action confidence evaluation (Section~\ref{self adaptive reward}).
The self-adaptive reward map can be considered a value map with the same size as the original input image (each agent has its self-adaptive reward), which can reflect the performance of the corresponding agent's action.
Besides, MECCA will suggest some low confidence regions for the user to interact next (Section~\ref{inter guid}).
Besides, the confidence map can also be used to generate the simulated label by comparing it with actions, which will be described in Section~\ref{unlabeled}.
The newly obtained hint map, the adjusted segmentation probability result, and the original 3D image form a new state.
The process described above is repeated until the segmentation result meets requirements.
To emphasize, during the testing stage in Figure~\ref{testing}, there is no need to calculate the self-adaptive rewards, and at the same time, we only need the obtained actions and suggested interaction areas.



\subsection{MARL-driven Interactive Segmentation Framework}
\label{MARL}
This study employs the multi-agent reinforcement learning structure to formulate the interactive segmentation process and continuously give error-corrective actions at each interaction step.
Let ${\mathbf{x}} = \left( x_1, \cdots, x_N \right)$ denotes the input image and $x_i$ denotes the $i$-th voxel of the image.
In the setting of MARL, every voxel $x_i$ is treated as an agent with its own  refinement policy $\pi_i({a_i}^{(t)},{s_i}^{(t)})$.
At time step $t$, agent $x_i$ gets action ${a_i}^{(t)}$ from the segmentation network according to its current state ${s_i}^{(t)}$. After taking the action, agent will receive a reward ${r_i}^{(t)}$ according to the segmentation result.

The state ${s_i}^{(t)}$ for agent $x_i$ is concentrated by its voxel value $b_i$, its current segmentation probability ${p_i}^{(t)}$ and the value ${h_i}^{(t)}$ on the hint map.
In particular, the segmentation probabilities of all agents are initialized to 0.5 and range from 0 to 1.
The hint map ${h}^{(t)}$ is transformed from the user’s hints which are in the form of edge points. 
At each step, users click on some edges, which are not correctly predicted, as hints. 
In order to let the model receive hint information, we generates a 3D Gaussian (with $8$-voxels kernel size) centered on each of the edge points as the hint map input to the segmentation network. 

The action ${a_i}^{(t)}$ for agent $x_i$ is sampled from its policy and used to adjust its previous segmentation probability:
\begin{equation}
a_i^{(t)}\sim\pi_{\theta}(a_i^{(t)}|s_i^{(t)}),
\end{equation}
\begin{equation}\label{eq:clip}
p_i^{(t+1)}= \ clip(p_i^{(t)}+a_i^{(t)},0,1),
\end{equation}
where ${a_i}^{(t)} \in \mathcal{A}$ and the $clip$ operation modifies the probability to the interval [0,1]. 
The action set $\mathcal{A}$ contains actions of different scales, allowing the agent to select the proper action. 
In our setting, the $\mathcal{A}= \{\pm0.1, \pm0.2, \pm0.4\}$.
$0$ is not used as one of the actions is mainly due to the following three reasons:
1) in the early stages of the experiment, we found that using $0$ as action will make the algorithm converge slow, and it requires more than $2$ times the number of interactions to achieve the performance of baselines.
Therefore, we remove $0$ from the action space and encourage the agent to explore in the early stage of training. 
Not only does the algorithm converge faster, but it also requires fewer interactions;
2) in order to prevent the algorithm's output from exceeding the valid range, we have clipped the output action (this is also a common technique in reinforcement learning in order to ensure that the output of the policy is in the valid range), as shown in the (\ref{eq:clip}).
3) due to the existence of the self-adaptive confidence calibration mechanism, the action output by the algorithm will generally change in the direction of the correct label. 
Combined with the clipping technique used in the second point, the algorithm can guarantee effectual output even if there is no $0$-action.

The reward ${r_i}^{(t)}$ is the feedback (positive or negative) of the action and used to update the refinement policy. 
The design of reward is a significant part of our algorithm, and we will introduce it in details in Section~\ref{self adaptive reward}.


\subsection{Segmentation Network}
\label{seg train}

\begin{figure*}[ht!]
\centering
\includegraphics[width=\textwidth]{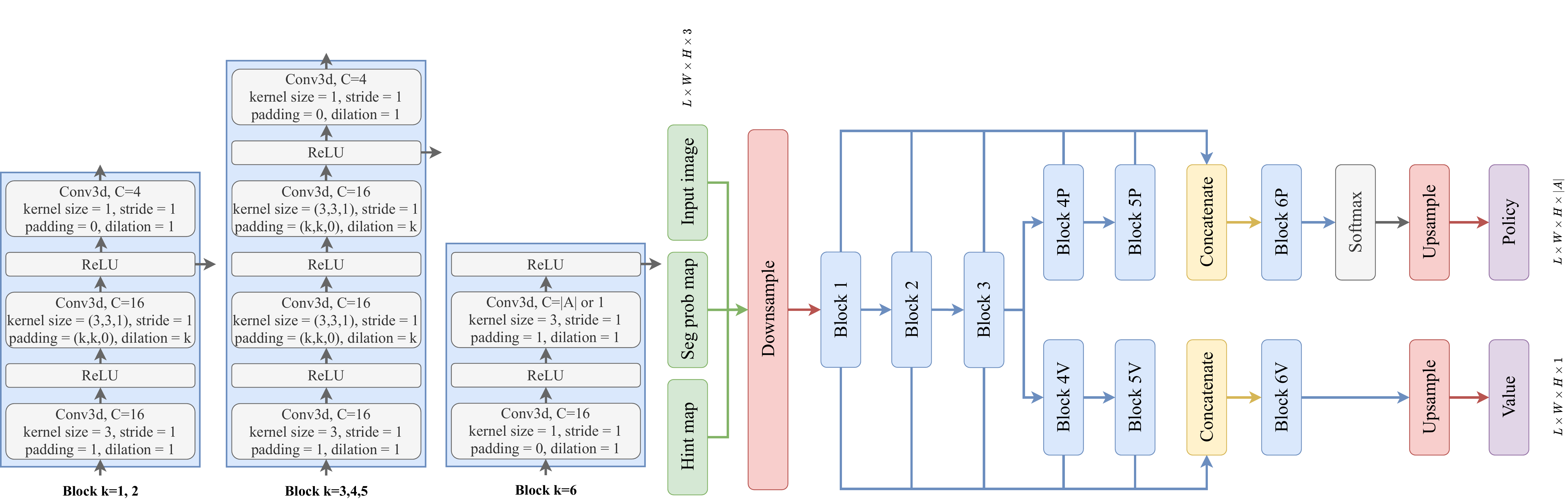}
\caption{{The detailed architecture of the segmentation network. The confidence network architecture is the same as the value branch of the segmentation network, but the parameters are not shared. $|A|$ represents the size of the action space.}}
\label{network}
\end{figure*}

The segmentation network adopts the P-Net, proposed in ~\cite{wang2018deepigeos}, as the backbone. 
The segmentation network $\bm{S}$ has two heads: policy head and value head. 
Two heads share the first three 3D convolutional blocks to extract low-level features.
Each of the blocks has two convolution layers and the size of the convolution kernel is fixed as 3×3 in all these convolution layers. 
All the convolution kernels are dilated convolution, which can reduce the loss of the resolution.
Both of the heads have another two 3D convolutional blocks to extract specific high-level features.
The detailed architecture is shown in Figure~\ref{network}.
Specifically, the policy head is to output the policy $\pi_i({a_i}^{(t)},{s_i}^{(t)})$, which is the distribution of action probabilities under current state. 
By taking actions on different scales, the probabilities will be dynamically adjusted. 
The value head is to estimate the value of the current state, which evaluates how good the current state is and estimates the expected return:
\begin{equation}
V_i^{(t)} = \mathbb{E} \left[ R_i^{(t)} \big| s_i=s_i^{(t)} \right] = \mathbb{E} \left[ {\textstyle{\sum_{k=t}^{\rm T}}} \gamma^{t-1} r_{i}^{(t)} \right],
\end{equation}
where $T$ is the terminal time step in the interaction process, and $\gamma$ is the discount factor. 
$\theta_v$ denotes the parameters of the value head, and the gradient with respect to $\theta_v$ is:
\begin{equation}
{\hbox{d}}_{\theta_{v}}=\nabla_{\theta_{v}} A^{2} \big( s^{(t)}, a^{(t)} \big),
\label{critic}
\end{equation}
\begin{equation}
A\big(s^{(t)}, a^{(t)}\big)= {\textstyle{\sum_{k=t}^{T}}} \gamma^{k-t} \bar{r}^{(k)}-V \big(s^{(t)}\big),
\end{equation}
where $\overline{r}^{(k)}$ is the mean reward of all voxels at timestep $k$. 
$A(\bm{s}^{(t)}, \bm{a}^{(t)})$ represents the advantage function~\cite{Mnih2016AsynchronousMF}. 

The goal of the policy head is to maximize the expected return by selecting proper actions in state $\bm{s}^{(t)}$. 
This study uses $\theta_p$ to denote the parameters of the policy head, and the gradient for $\theta_p$ is denoted as
\begin{equation}
{\hbox{d}}_{\theta_{p}} = -\nabla_{\theta_{p}} \pi\big(a^{(t)} | s^{(t)}\big) A\big(a^{(t)}, s^{(t)}\big).
\label{actor}
\end{equation}
Usually, the policy head is updated after the value head.

\subsection{Action Confidence Learning}
\label{action confidence}
As we mentioned in section \ref{introduction}, there will be some situations where the segmentation model misunderstands or ignores the hint information. 
To some extent, these samples (or regions) with the phenomenon of interactive misunderstanding are hard samples (or regions). 
Although these samples may account for a small percentage of the dataset, they are critical for improving generalization and robustness. 
The most important thing is finding a \textit{professional easy-or-hard representer}~\cite{nie2019difficulty} to identify them when interacting. 
Focal loss~\cite{lin2017focal} evaluates the hard-or-easy samples through predicted probability. 
~\cite{nie2019difficulty} applies adversarial learning to train the easy-or-hard representer. 
Both of them have their advantages, but all evaluate hard-or-easy samples based on the final segmentation result. 
As such, these methods cannot be directly applied to interactive segmentation. 
For example, if the model predicts the category probability of a voxel to be $0.8$ ($p_i^{(t)}$ = $0.8$), and then takes an action $a_i^{(t)}$ = $-0.1$. 
The next prediction will be $0.7$ after interaction ($p_i^{(t+1)}$ = $0.7$). 
If $y_i=1$, before and after results are all correctly predicted because $p_i^{(t)}$ and $p_i^{(t+1)}$ are all greater than 0.5. 
While for interactive segmentation, the probability is changing in the wrong direction.
This change is what we called the \textit{interactive misunderstanding} phenomenon and the formally definition is shown in following.

\begin{definition}[\textit{Interactive Misunderstanding}]
For a binary classification problem, the sign of the foreground label $y=1$ is denoted as \textit{positive}, and the sign of the background label $y=0$ is denoted as \textit{negative} accordingly.
In an interactive medical image segmentation task (i.e. a voxel-wise binary classification problem), for any voxel $i$, if the sign of the change of segmentation probability output by algorithm for two consecutive interaction steps \textit{sign($\triangle(p^{(i)})$)} is not equal to \textit{sign($y_i$)}, then this phenomenon is defined as \textit{interactive misunderstanding}.
\end{definition}

Our proposed confidence network learns the confidence of given actions to avoid misunderstanding hint information and take accurate actions. 
We argue that confidential information can be used to regularized action choices and suggest more efficient interaction. 
The confidence network structure also uses P-Net ~\cite{wang2018deepigeos} as the backbone.
The confidence network contains six 3D convolutional blocks. 
Each of the blocks has two convolution layers, and the size of the convolution kernel is fixed as 3×3 in all these convolution layers. 
The detailed architecture is shown in Figure~\ref{network}.

The confidence network is trained using the previous state and action as input and a confidence map as output. 
The confidence network is optimized by minimizing the summation of binary cross-entropy loss over actions (shown in (\ref{CLoss})) at each time step $t$. 
Here we use $\bm{C}$ to denote the confidence network, $w_{\bm{C}}$ denotes the parameters of the confidence network, while $L_{BCE}$ denotes the binary cross-entropy loss:
\begin{equation}
\begin{aligned}
&L_{\bm{C}}(s^{(t)}, a^{(t)} ; w_{\bm{C}})\\
=& L_{B C E}(\bm{C}(s^{(t)}, a^{(t)}), \operatorname{g}^{(t)}) + L_{B C E}(\bm{C}(s^{(t)},-a^{(t)}), 1-\operatorname{g}^{(t)}),
\end{aligned}
\label{CLoss}
\end{equation}
where
\begin{equation}
\operatorname{g}^{(t)}=\left\{\begin{array}{ll}
0 & \text { if } a^{(t)} \oplus y^{(t)}==1, \\
1 & \text { otherwise },
\end{array}\right.
\end{equation}
where $\operatorname{g}^{(t)}$ means whether the direction of action is consistent with the label. $a\oplus b$ is defined as that the statement is only true if either $a>0$ or $b>0$, but not both. 

One potential issue when training the confidence network is the imbalance of samples. 
Early in RL training, the error rate of actions output by the segmentation network is very high, while most actions are correct when the network gradually converges. 
Inspired by the training of the discriminator in generative adversarial networks, this study introduces symmetric samples into the (\ref{CLoss}) to speed up training. 
An obvious advantage is that it improves sample utilization efficiency because the confidence network can know what a ``bad sample" is and get a corresponding ``good sample."

\subsection{Self-Adaptive Reward}
\label{self adaptive reward}
Essentially, no matter applying focal loss or adversarial learning to the training of the segmentation model, they all try to weigh these hard samples to prevent the model dominated by easy samples. 
However, the interactive segmentation task differs from those fully automatic segmentation tasks because the interactive segmentation model needs to cooperate with the user and understand the user's hint information. 
The action to refine the segmentation result shows how the segmentation model understands hint information. 
Therefore, it is necessary to ensure that the hint information is correctly understood and the correct action is taken.

Specifically, the previously described action-confidence learning can provide the segmentation model with a confidence map to alleviate the interactive misunderstanding phenomenon. 
By this confidence map, hard-or-easy samples can be better recognized as the confidence values for these 'hard regions' are relatively lower than in other regions. 
This paper formulates this voxel-level action-aware as the self-adaptive reward function, $r^{(t)}$, which is shown in (\ref{action-aware}), to adapt this mechanism to the training of MARL:
\begin{equation}
\begin{aligned}
r^{(t)} = {\textstyle{\sum_{i=1}^{N}}} \alpha\left(2-c_{i}\right)^{\beta}{gain}_i^{(t)},
\end{aligned}
\label{action-aware}
\end{equation}
where $c_{i}$ is the value on the confidence map. 
The setting of hyperparameters $\alpha$ and $\beta$ are described in Section \ref{details}. 
In addition, ${gain}_i^{(t)}$ denotes the relative gain of cross-entropy:
\begin{align} 
\label{reward}
{gain}_i^{(t)} \ = \mathcal{X}_{i}^{(t-1)} - \mathcal{X}_{i}^{(t)},
\end{align}
\begin{align} 
\label{cross entropy}
\mathcal{X}_{i}^{(t)} \ =\ -y_{i}\log(p_{i}^{(t)})-(1-y_{i})\log(1-p_{i}^{(t)}),
\end{align}
where $\mathcal{X}_{i}^{(t)}$ denotes the cross-entropy between current segmentation probability and ground truth. 
If an agent gets a positive reward, its current action is good, and the refined segmentation result is closer to the ground truth.

With the self-adaptive reward function in (\ref{action-aware}), the confidence value $c_i$ of these wrong actions is lower, and they will be punished more when training our segmentation model.

\subsection{Interaction Guide}
\label{inter guid}
\begin{figure}[ht!]
\centering
\includegraphics[width=0.8\textwidth]{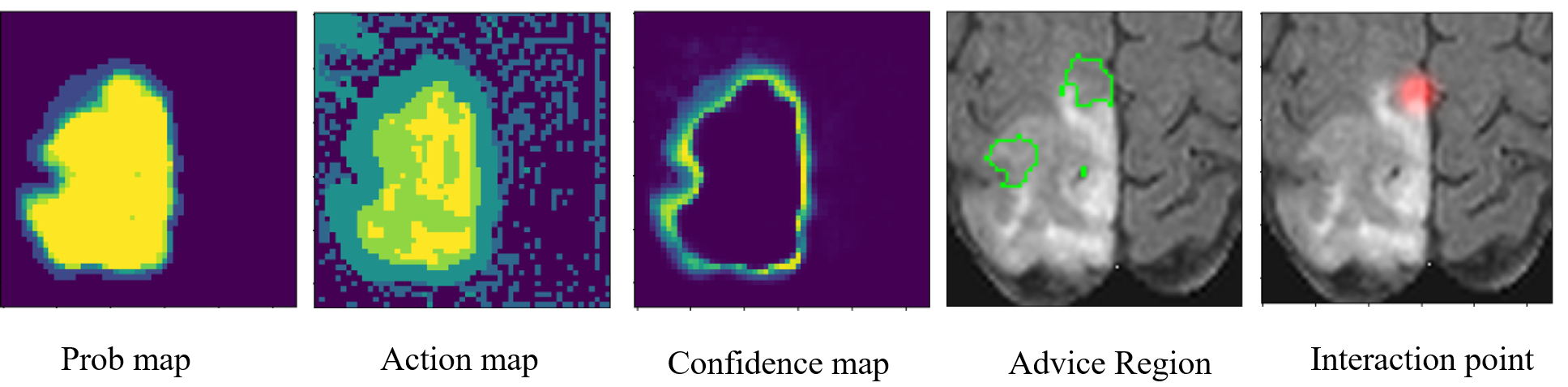}
\caption{The illustration of interaction guide mechanism. The areas surrounded by the green lines are advice regions provided to users, and the red point is the real hint information selected from advice regions. The brighter the color (closer to yellow), the larger the positive value; on the contrary, the darker the color (closer to black), the smaller the negative value.}
\label{InterGuid}
\end{figure}

Another challenge for interactive image segmentation is that users usually need to decide where to interact in many slices, which is very time-consuming, especially for 3D images. 
Our framework also provides users with an interaction guide mechanism to save the user's interaction time. 
After refinement, our framework will suggest some possible areas for users to interact with next. 
Specifically, our framework will filter out those areas with low action confidence and provides them for users (see Figure~\ref{InterGuid}).
Firstly, the original 3D image will be segmented with super voxels;
each super voxel can be regarded as a group of voxels that share common characteristics.
We use simple linear iterative clustering (SLIC)~\cite{Achanta2012SLICSC} technique with \textit{spacing} $=[2,2,2]$, \textit{compactness} $=0.1$ to generate supervoxels, and the number of initial supervoxels equals to $100$ and gradually declines during the refinement iterations for training and testing.
Secondly, the proposed algorithm will compute the mean action confidence in each super voxel and rank them in descending order. 
Finally, the top $5$ super voxels will be marked and recommended to users. 
What users need to do is to select the best interaction positions from these super voxels. 

\subsection{Simulated Label Generation}
\label{unlabeled}

\begin{figure}[ht!]
\centering
\includegraphics[width=0.8\textwidth]{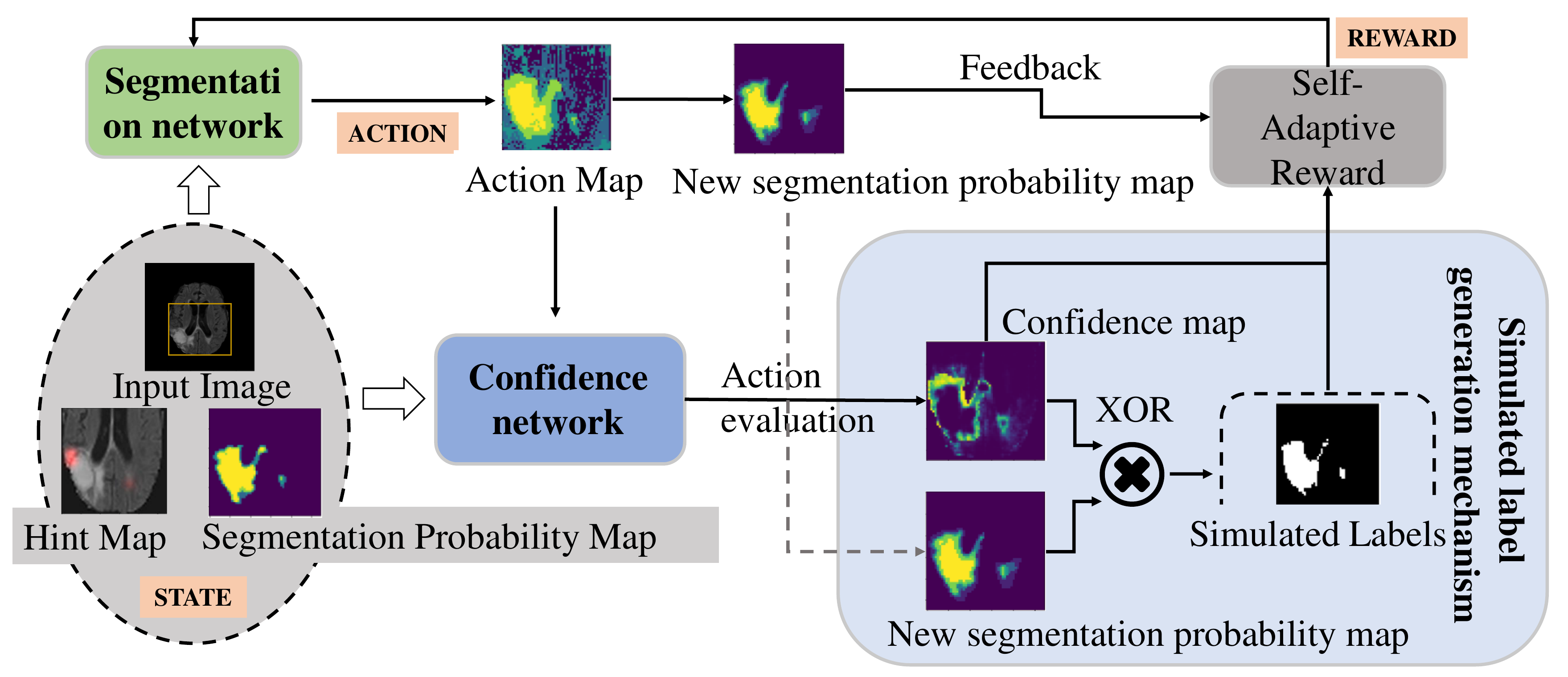}
\caption{The illustration of simulated label generation. The simulated label generation mechanism utilizes the confidence map and the action map to generate the simulated label. The confidence map is used to calibrate the action, and the direction of the calibrated action is the simulated label of each voxel.}
\label{labelgen}
\end{figure}

In medical imaging, the unlabeled data is much more than labeled data due to difficulties in labeling medical images. 
To address this lack of annotations, the proposed algorithm leverages the action confidence not only to improve the utilization efficiency of hint information but also to generate a simulated label, shown in Figure \ref{labelgen}, for unlabeled data. 
We define the simulated label as follow:
\begin{equation}
\label{simulated label}
\hat{y}^{(t)}=\left\{\begin{array}{ll}
1 & \text { if } a^{(t)} \oplus c^{(t)}==1, \\
0 & \text { otherwise }.
\end{array}\right.
\end{equation}
The $\hat{y}^{(t)}$ is the simulated voxel-level label generated from the confidence map and detailed illustration is shown in Figure~\ref{labelgen}.
Based on this mechanism, this study considers using these unlabeled data to assist in training the segmentation network. 
Specifically, this study divides the dataset into two parts during the training stage: the first part contains voxel-level annotation information, and the second part contains only hint information during training. 
The labeled data is trained as above, while the unlabeled data is trained with the simulated label. During the training stage, the backward of gradients will be masked: 
\begin{equation}
\begin{aligned}
d \theta_{v}=M\nabla_{\theta_{v}} A\big(s^{(t)}, a^{(t)}\big)^{2},
\end{aligned}
\label{fake_v}
\end{equation}

\begin{equation}
\begin{aligned}
d \theta_{p}=M\nabla_\pi \big(a^{(t)} | s^{(t)} \big) A \big(a^{(t)}, s^{(t)} \big),
\end{aligned}
\label{fake_p}
\end{equation}
where 
\begin{equation}
\begin{aligned}
M = I(max(c^{(t)},1-c^{(t)})>\delta).
\end{aligned}
\label{mask}
\end{equation}
$A\left(\boldsymbol{s}^{(t)}, \boldsymbol{a}^{(t)}\right)$ is the advantage (which is defined in~\cite{Mnih2016AsynchronousMF}) at time step $t$ of taking $a^{(t)}$ in condition of state $s^{(t)}$, which indicates the actual accumulated reward without being affected by the state and reduces the variance of gradient.
A mask, $M$, is used to constraint the training of unlabeled data. 
The gradients of unlabeled data backward only when the action confidence exceeds the threshold $\delta$ (which is gradually increased during the training process). 
Unlike traditional pseudo-label training, the supervised signal does not come from the segmentation but the confidence network.
These filtered data with hint information are more valuable and provide more accurate supervised signals.
Generally, the training process with labeled and simulated labeled data is carried out simultaneously. 
When using the simulated label generation mechanism, the pseudocode is shown in {\bf{Algorithm~\ref{alg:1}}}.

\begin{algorithm}
	\caption{MECCA: Interactive Medical Image Segmentation with Self-Adaptive Confidence Calibration}
	\label{alg:1}
	\begin{algorithmic}[1]
		\STATE Initialize the segmentation network $\bm{S}$ with $\theta$;
		\STATE Initialize the confidence network $\bm{C}$ with $w$;
		\FOR{every sample in labeled datasets}
		\STATE Set the segmentation probability of each voxel to $0.5$;
		\STATE $s^{(0)} \leftarrow (x,p^{(0)},h^{(0)})$;
		\FOR{every interaction time step $t$}
		\STATE Take action $a^{(t)} \leftarrow S(s^{(t)})$;
		\STATE Get reward $r^{(t)}$ and observe the next state $s^{(t+1)}$;
		\STATE Compute the gradient of $\bm{S}$ via (\ref{critic}) and (\ref{actor});
		\STATE Compute the gradient of $\bm{C}$ via (\ref{CLoss});
		\ENDFOR
		\STATE Get a sample ($x', y'$) from unlabeled dataset;
		\STATE Initialize the state: $s'^{(0)} \leftarrow (x',p'^{(0)},h'^{(0)})$;
		\FOR{every interaction time step $t$}
		\STATE Take action $a'^{(t)} \leftarrow S(s'^{(t)})$;
		\STATE Generate the simulated label $\hat{y}^{(t)}$ by (\ref{simulated label});
		\STATE Observe the reward and next state $s'^{(t+1)}$;
		\STATE Compute the gradient of $\bm{S}$ via (\ref{fake_v}) and (\ref{fake_p}).
		\ENDFOR
		\ENDFOR
	\end{algorithmic}  
\end{algorithm}


\section{Experiments and results}
\label{exp}

\subsection{Dataset and Implementaion details}\label{details}
To comprehensively evaluate our proposed method, we apply our algorithm to four 3D medical image datasets. 
All datasets are divided into two parts: $D_{train}$/$D_{test}$.
The details of these datasets are as follows: 
1) \textbf{BraTS2015}: Brain Tumor Segmentation Challenge $2015$ ~\cite{menze:hal-00935640} contains $274(234/40)$ multiparametric MRI(Flair, T1, T1C, T2) from brain tumor patients. 
In our task, we only use the Flair image and segment the whole brain tumor. 
2) \textbf{BraTS2020}: Brain Tumor Segmentation Challenge $2020$ ~\cite{menze:hal-00935640} contains $285(235/50)$ multiparametric MRI(Flair, T1, T1C, T2) from brain tumor patients. 
In our task, we only use the Flair image and segment the whole brain tumor. 
3) \textbf{MM-WHS}: Multi-Modality Whole Heart Segmentation~\cite{zhuang2016multi} contains $24(20/4)$ multi-modality whole heart images covering the whole heart substructures. 
In our task, this study chooses to segment the left atrium blood cavity. 
4) \textbf{Medical Segmentation Decathlon}: This is a generalisable 3D semantic segmentation datasets containing different organ segmentation tasks~\cite{simpson2019large}. 
This study chooses to use the spleen and liver dataset, which provides $61(41/20)$ and $106(96/10)$ CT images respectively.  

We implement our method with PyTorch~\cite{NEURIPS2019_9015}\footnote{The demo video of MECCA algorithm is available at \url{https://bit.ly/mecca-demo-video}.}. 
The segmentation and confidence network are both initialized by Xavier~\cite{glorot2010understanding} method, and learning rates are initialized to $1e-4$. 
Other parameters are set as follow: $T=5$, $\gamma=0.95$, $\alpha=0.8$, $\beta=1$. 
The mask $M$ ranges from $0.85$ to $0.99$ and ascents $0.00025$ at every epoch. 
Adam~\cite{kingma2014adam} is adopted as the optimizer. 
The original image is cropped by the bounding box based on the ground truth with a random extension in the range of $1$ to $11$ voxels. 
Each image is then resized and normalized to $55 * 55 * 30$.
The data is augmented by random flip and rotation.
The proposed algorithm training time with one Nvidia 2080ti GPU varies from $5$ to $13$ hours for different datasets.

\subsection{Interaction Settings and Evaluation Metrics}

\begin{figure}[ht!]
\centering
\includegraphics[width=0.6\textwidth]{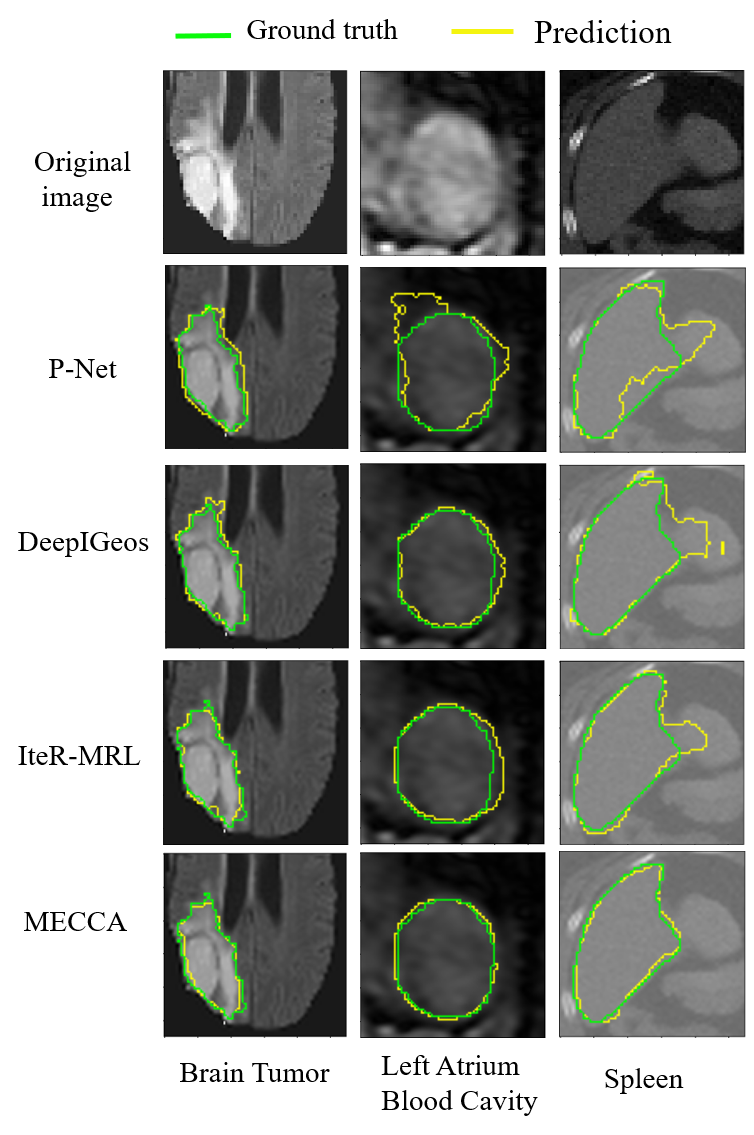}
\caption{Visualization of segmentation results by different methods. The green lines represent the boundaries of ground truth, and the yellow lines represent the predictive boundaries.}
\label{view1}
\end{figure}

During the process of interaction, this study adopts the edge points as the hint information for two reasons. 
On the one hand, click operation saves more time than scribbles or other interactive ways. 
On the other hand, edge points can provide more information about object edges as the bottleneck of medical image segmentation is usually the inaccurate segmentation of object edges. 
Specifically, this study provides each method with $45$ points\footnote{Please refer to Section IV-C for details.} during the whole interaction process. 
At every step, users will click some edges which are not correctly predicted. 
This study generates a 3D Gaussian (with $8$-voxels kernel size) centered on each of the edge points as the hint map to let networks receive hint information. Then the hint map will be input to the segmentation network as part of the state. 
This study will use the Dice score and the average symmetric surface distance (ASSD) to evaluate the performance of the segmentation result. According to these evaluation metrics, doctors can judge the patient's condition:
$$
{\text{Dice}}(S_p, S_g) = (2|S_p \cap S_g|) / (|S_p| + |S_g|),
$$
where $S_p,S_g$ denote the prediction of an algorithm and the ground truth respectively.
$$
{\text{ASSD}} = \big({\textstyle{\sum_{i \in \mathcal{S}_{a}}}} d\left(i, \mathcal{S}_{b}\right) + {\textstyle{\sum_{i \in \mathcal{S}_{b}}}} d\left(i, \mathcal{S}_{a}\right)\big) / (\left|\mathcal{S}_{a}\right|+\left|\mathcal{S}_{b}\right|),
$$where $\mathcal{S}_{a}$ and $\mathcal{S}_{b}$ represent the set of surface points of the segmentation result predicted by the algorithm and the ground truth, respectively. 
$d\left(i, \mathcal{S}_{b}\right)$ is the shortest Euclidean distance between $i$ and $\mathcal{S}_{b}$. 
The Dice score and ASSD in all tables are the average value of five test results of algorithms.

\subsection{Comparisons with State-of-the-art Methods}

\begin{table}
\caption{Dice scores at each interaction step by different methods. The value in parentheses represents the improvement relative to the previous step.}
\label{improvementTab}
\centering
\scalebox{0.9}{
\begin{tabular}{c c c c c c c} 
\toprule[1pt]
Step  & 1 & 2 & 3 & 4 & 5 \\
\hline
\multirow{2}{*}{DeepIGeos} & 87.36&87.53&87.67&87.99&88.32 \\
                    & & (+0.17) & (+0.14) & (\textbf{+0.32}) & (+0.33) \\
\hline
\multirow{2}{*}{InterCNN} & 87.21 & \textbf{88.59} & 88.54 & 88.39 & 88.26 \\
 & & (+1.38) & (-0.04) & (-0.16) & (-0.12) \\
\hline
\multirow{2}{*}{IteR-MRL} & 84.56 & 85.35 & 88.15 & 88.11 & 88.94 \\
 & & (+0.80) & (\textbf{+2.80}) & (-0.08) & (+0.83) \\
\hline
\multirow{2}{*}{MECCA} & 86.49 & 88.53 & \textbf{89.56} & \textbf{89.12} & \textbf{90.29} \\ 
 & & (\textbf{+2.03}) & (+1.03) & (-0.44) & (\textbf{+1.17})  \\[0.5ex] 
\bottomrule[1pt]
\end{tabular}}
\end{table}

\begin{figure}[ht!]
\centering
\includegraphics[width=0.6\textwidth]{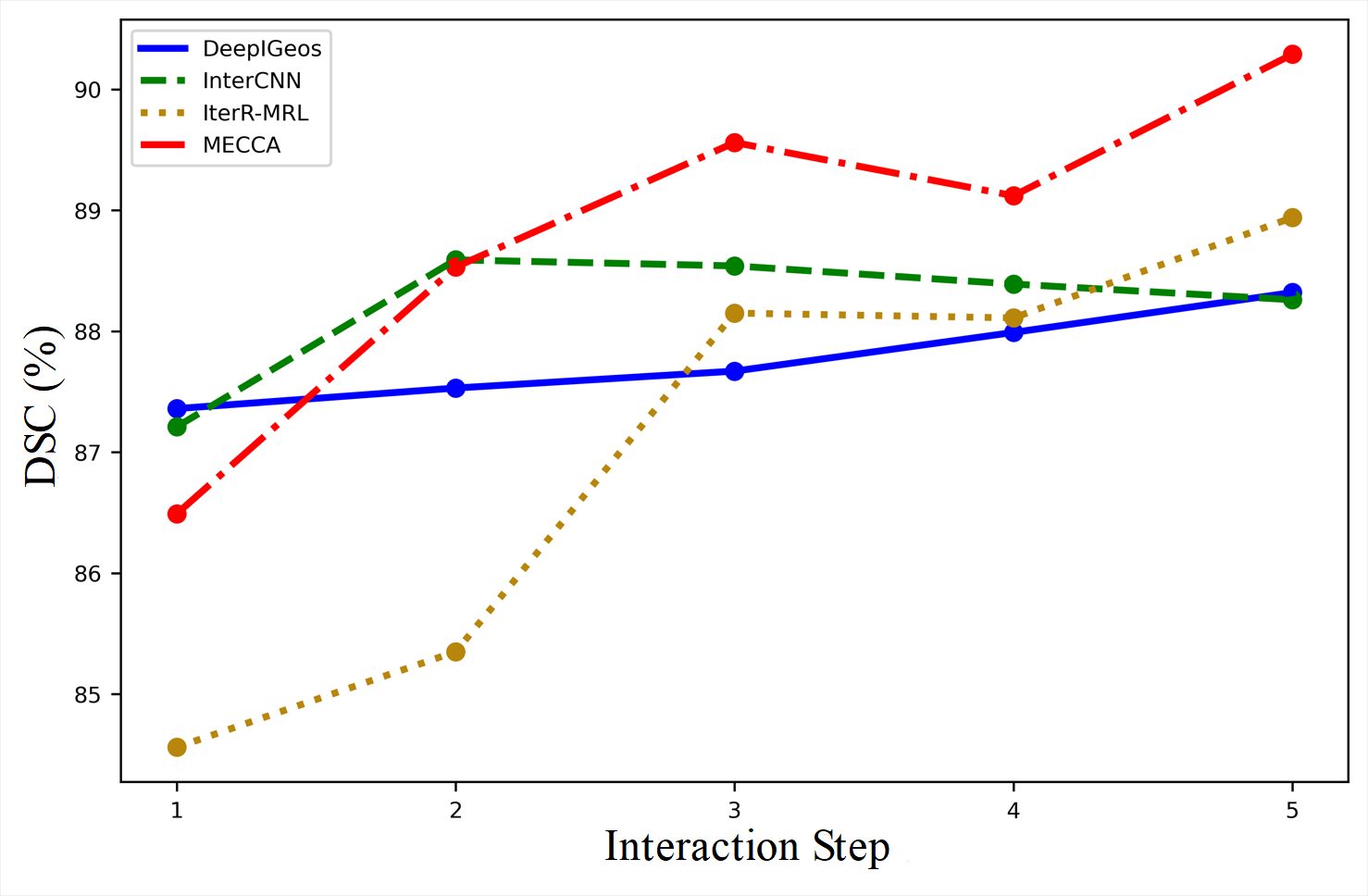}
\caption{ Visualization of the performance improvement of different interactive medical segmentation methods at the different interaction steps. All these testing results are performed on the BraTS2015 dataset.}
\label{improvementFig}
\end{figure}

We compare MECCA with five state-of-the-art interactive segmentation methods: DeepIGeos~\cite{wang2018deepigeos}\footnote{\url{https://github.com/taigw/GeodisTK.}}, InterCNN~\cite{bredell2018iterative}\footnote{\url{https://github.com/gbredell/interCNN.}}, IteR-MRL~\cite{liao2020iteratively} and
BS-IRIS~\cite{ma2020boundary-aware}. 
InterCNN is the multi-step version of DeepIGeos.
We also introduce the SOTA method, U-Net~\cite{ronneberger2015u}\footnote{\url{https://github.com/liyun-lu/unet\_and\_vnet.}}, of medical image segmentation as a comparable baseline.
Table~\ref{res1} show the quantitative comparison of the six interactive segmentation methods on different datasets. 
For a fair comparison, all CNN-based methods adopt the same network structure (P-Net), which is proposed in~\cite{wang2018deepigeos}.
We can see that our proposed MECCA performs better than other state-of-the-art methods on all three datasets. 
This study also visualizes the results in Figure~\ref{view1}, which shows that our method is more accurate in edge segmentation.

\begin{table*}[htbp]
\label{tab:test}
\caption{Quantitative comparison of 3D medical image datasets segmentation by different methods. In particular, the P-Net is the method without hint information. Significant improvement
(p-value $< 0.05$) is marked in bold.}
\label{res1}
\centering
\resizebox{\textwidth}{!}{%
 \begin{tabular}{c|cccccccccc}
\toprule[1pt]
Methods &\multicolumn{2}{c}{BraTS2020} &\multicolumn{2}{c}{BraTS2015} & \multicolumn{2}{c}{MM-WHS} & \multicolumn{2}{c}{Spleen} & \multicolumn{2}{c}{Liver}\\
        & Dice(\%) & ASSD(pixels) & Dice(\%) & ASSD(pixels) & Dice(\%) & ASSD(pixels) &Dice(\%) & ASSD(pixels) &Dice(\%) & ASSD(pixels)\\
\midrule
P-Net            &83.67$\pm$8.35 &5.78$\pm$4.01 &84.00$\pm$12.01  &5.10$\pm3.72$ &81.40$\pm$1.48  &3.28$\pm$0.45 &88.08$\pm$2.25 &4.25$\pm$2.07 &35.89$\pm$2.61 &34.46$\pm$23.82 \\

U-Net            &84.72$\pm$10.42 &4.09$\pm$3.89 &84.66$\pm$11.25 &6.17$\pm$4.69 &80.96$\pm$1.65 &3.72$\pm$0.39 &87.95$\pm$2.87 &5.12$\pm$1.09  &56.00$\pm$1.93 &22.38$\pm$22.42\\

DeepIGeos &88.54$\pm$.97 &2.11$\pm$1.30 &88.32$\pm$5.34  &2.28$\pm$1.24  &88.48$\pm$0.71  &1.53$\pm$0.18 &91.97$\pm$1.51 &0.93$\pm$0.46 &48.57$\pm$2.52 & 10.28$\pm$3.45\\

InterCNN        &88.39$\pm$6.01 &2.01$\pm$1.09 &88.26$\pm$7.07  &1.81$\pm$2.09  &87.85$\pm$1.15  &0.80$\pm$0.15 &93.52$\pm$0.94 &0.54$\pm$0.83 &59.92$\pm$2.20 &5.95$\pm$2.76\\

IteR-MRL        &89.22$\pm$4.65 &2.07$\pm$0.91 &88.94$\pm$4.81   &\textbf{1.41$\pm$0.22}  &89.55$\pm$0.87  &0.90$\pm$0.11 &91.50$\pm$1.34 &0.67$\pm$0.21 &62.29$\pm$1.93 &\textbf{0.87$\pm$0.59}\\

BS-IRIS        &90.47$\pm$5.23 &1.82$\pm$0.33 &89.74$\pm$3.86   &1.61$\pm$0.42  &89.12$\pm$0.98  &1.19$\pm$0.16 &92.35$\pm$1.13 &0.54$\pm$0.19 & 67.25$\pm$2.01& 4.34$\pm$1.18\\

MECCA        &\textbf{91.02$\pm$5.86} &\textbf{1.15$\pm$0.20} &\textbf{90.29$\pm$5.07}  &1.50$\pm$0.33  &\textbf{90.39$\pm$5.89} &\textbf{0.80$\pm$0.01} &\textbf{94.96$\pm$1.44} &\textbf{0.30$\pm$0.16} &\textbf{71.46$\pm$1.41} &2.36$\pm$0.99\\
\bottomrule
\end{tabular}%
}
\end{table*}


To demonstrate that our method can take advantage of hint information more efficiently, this study also compares the relative improvement of different methods at each interaction step. 
Due to the different interaction ways of different methods, this study sets up a unified interaction process: 
all these methods have $5$ interaction steps and receive $25$ identical points at the first step to generate the initial segmentation result. 
After that, users will click $5$ points at each step in the following $4$ interaction steps (total of $25 + 5\times 4 = 45$ points), and these methods will iteratively refine previous segmentation results. 
The DeepIGeos method does not model the interaction sequence, and it just combines current and previous hint information to refine the previous segmentation. 
The experimental results are shown in Table~\ref{improvementTab} and Figure~\ref{improvementFig}. 

The results show that MECCA can perform better under the same amount of hint information and improve more dice scores at most steps. 
Compared with CNN-based methods, the main advantage of RL-based methods is that they can always keep notable improvement. 
There are two reasons for this result. 
The first one is that RL-based methods model the whole interaction process to avoid interaction conflict. 
The second reason is the relative entropy-based reward which encourages the model to keep refining results. 
However, we should realize, the RL-based methods still can not guarantee the high confidence of the corrective actions. 
As we can see in Figure~\ref{improvementFig}, the performance of IteR-MRL is not as good as other methods at the beginning, which is caused by numerous incorrect actions. 
However, after learning the confidence of actions and applying the self-adaptive reward to update the segmentation model, our proposed MECCA can perform well at each step and keep significantly refining the result.

Figure~\ref{segproc} presents the visualization of the segmentation process of our proposed method. 
Since our proposed method models the whole interaction process, Figure~\ref{segproc} shows the current interaction step and its previous and next interaction steps. 
At each step, the second column is the confidence map. 
This study finds that the confidence value of object edges is always lower than in other regions, and these regions will receive more `punishment' when rewards are generated. 
This study observes that MECCA can gradually correct the edges around the user clicks (the red regions).

\begin{table}[ht!]
\centering
\caption{The DICE of our method varies with the number of interactions under different difficult cases.}
\label{tab:difficults-vs-interacts}
\resizebox{0.5\textwidth}{!}{%
\begin{tabular}{|c|l|l|l|l|l|}
\hline
\rowcolor[HTML]{9B9B9B} 
\multicolumn{1}{|l|}{\cellcolor[HTML]{9B9B9B}} & \multicolumn{5}{c|}{\cellcolor[HTML]{9B9B9B}Interaction times} \\ \hline
\rowcolor[HTML]{FFFFFF} 
 &
  \multicolumn{1}{c|}{\cellcolor[HTML]{FFFFFF}1} &
  \multicolumn{1}{c|}{\cellcolor[HTML]{FFFFFF}2} &
  \multicolumn{1}{c|}{\cellcolor[HTML]{FFFFFF}3} &
  \multicolumn{1}{c|}{\cellcolor[HTML]{FFFFFF}4} &
  \multicolumn{1}{c|}{\cellcolor[HTML]{FFFFFF}5} \\ \hline
\rowcolor[HTML]{FFFFFF} 
The easy case                                      & 94.05      & 95.15      & 95.38      & 95.91      & 96.25      \\ \hline
\rowcolor[HTML]{FFFFFF} 
The hard case                                       & 41.45      & 47.38      & 65.78      & 77.79      & 79.37      \\ \hline
\end{tabular}%
}
\end{table}

\begin{table}[htb!]
\centering
\caption{MECCA's tolerance for inaccurate interaction points. Significant improvement
(p-value $< 0.05$) is marked in bold.}
\label{tab:disturbed}
\resizebox{0.5\textwidth}{!}{%
\begin{tabular}{|c|l|l|}
\hline
& DICE (\%) & ASSD (pixels) \\ \hline
Disturbed Interactions & 88.75±9.72     & \textbf{1.11±0.23}     \\ \hline
Non-Disturbed Interactions                    & \textbf{90.29±5.07}     & 1.50±0.33     \\ \hline
\end{tabular}%
}
\end{table}

\begin{figure}[ht!]
\centering
\subfigure[DeepIGeos]{
\includegraphics[width=0.8\textwidth]{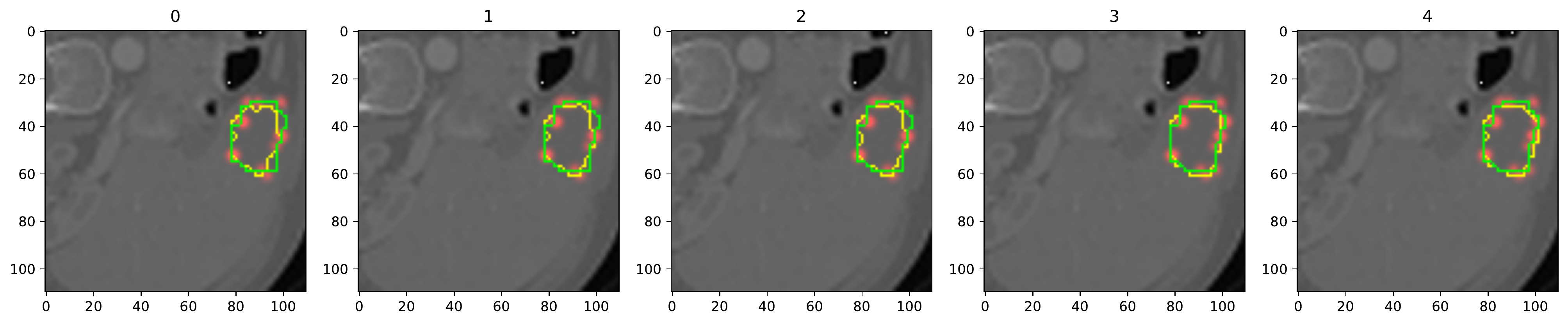}
}
\subfigure[InterCNN]{
\includegraphics[width=0.8\textwidth]{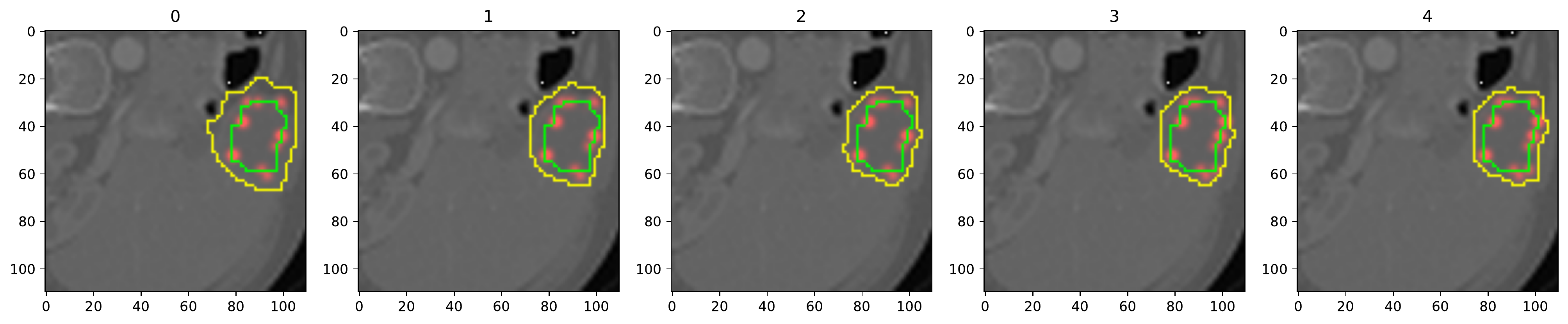}}
\subfigure[IteR-MRL]{
\includegraphics[width=0.8\textwidth]{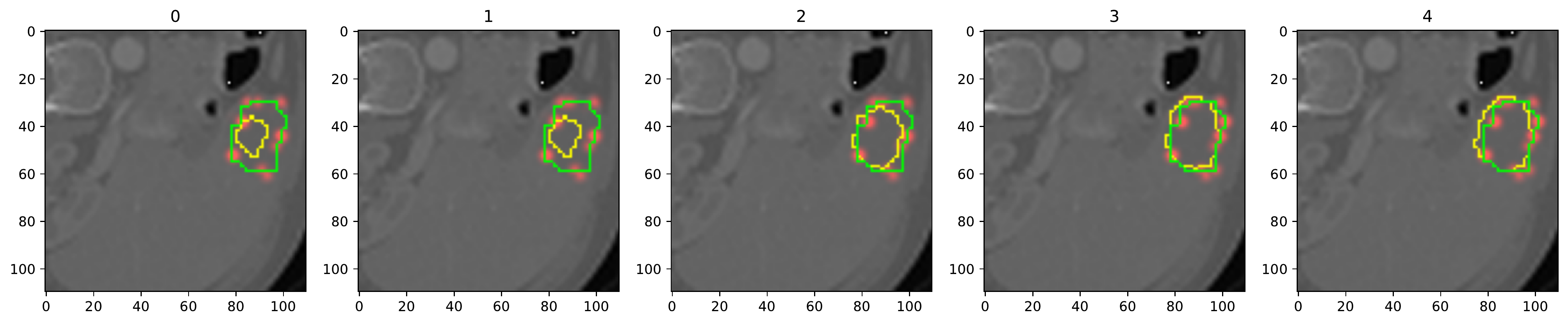}}
\subfigure[MECCA]{
\includegraphics[width=0.8\textwidth]{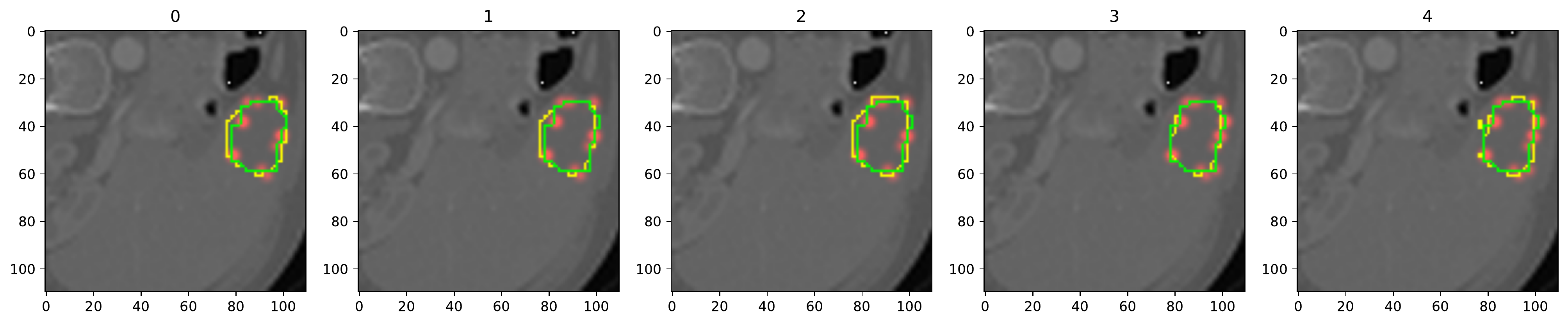}}
\caption{The results of different methods response to the same user interactions according to the same initial segmentation on 4/10 testing instance, 7/30 channel, Liver 2 of MSD dataset.}
\label{fig:sint1}
\end{figure}

\begin{figure}[ht!]
\centering
\subfigure[DeepIGeos]{
\includegraphics[width=0.8\textwidth]{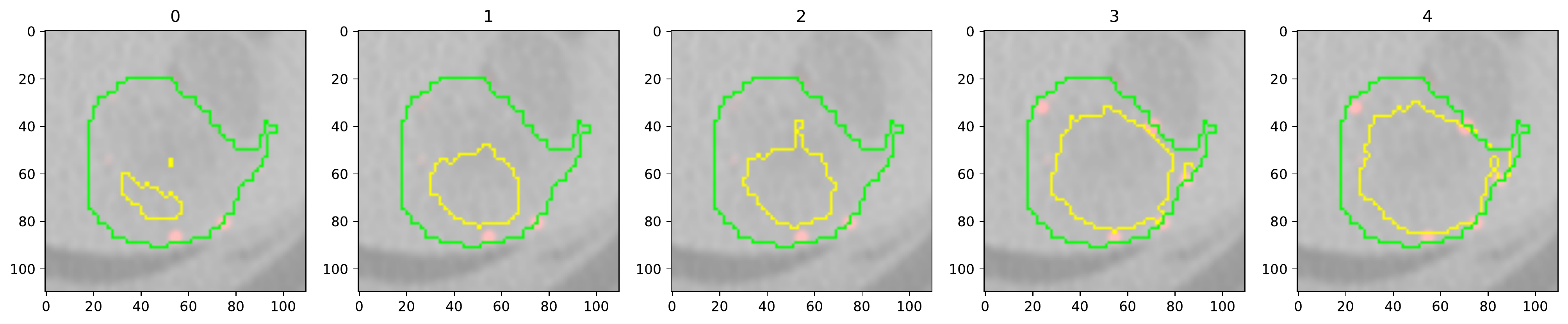}
}
\subfigure[InterCNN]{
\includegraphics[width=0.8\textwidth]{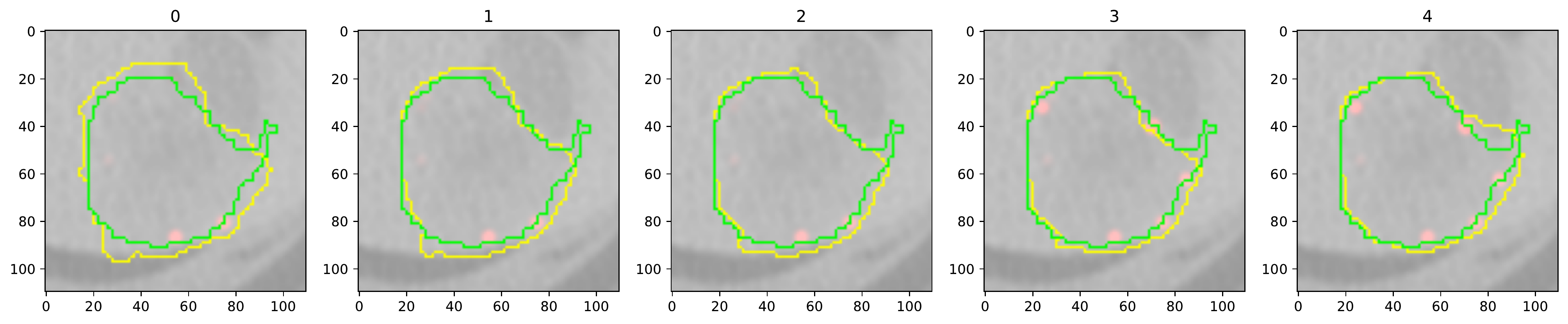}}
\subfigure[IteR-MRL]{
\includegraphics[width=0.8\textwidth]{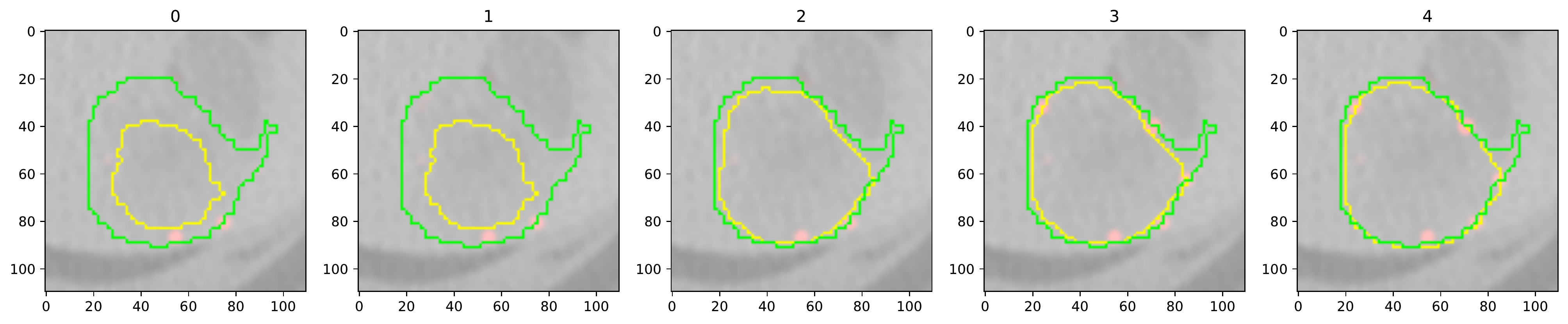}}
\subfigure[MECCA]{
\includegraphics[width=0.8\textwidth]{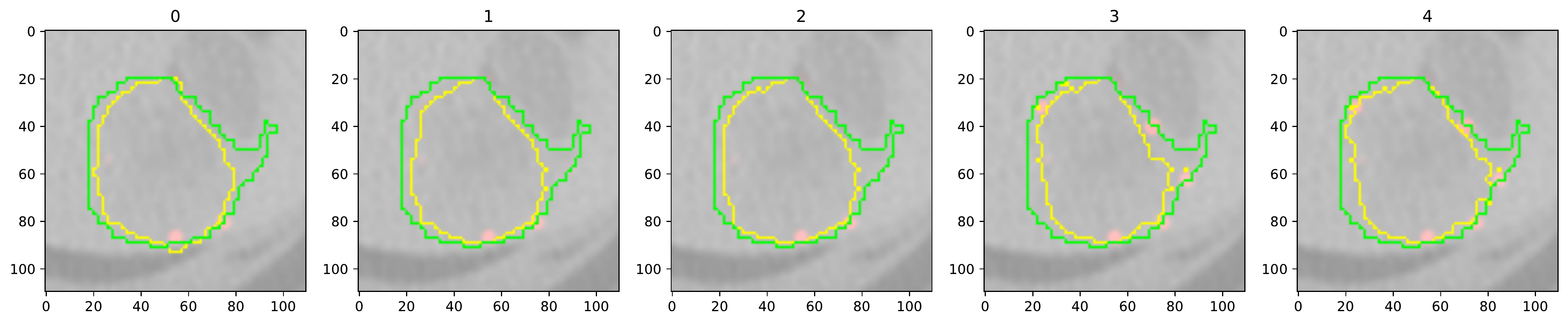}}
\caption{The results of different methods response to the same user interactions according to the same initial segmentation on 6/10 testing instance, 18/30 channel, Liver 2 of MSD dataset.}
\label{fig:sint2}
\end{figure}

\begin{figure}[ht!]
\centering
\subfigure[DeepIGeos]{
\includegraphics[width=0.8\textwidth]{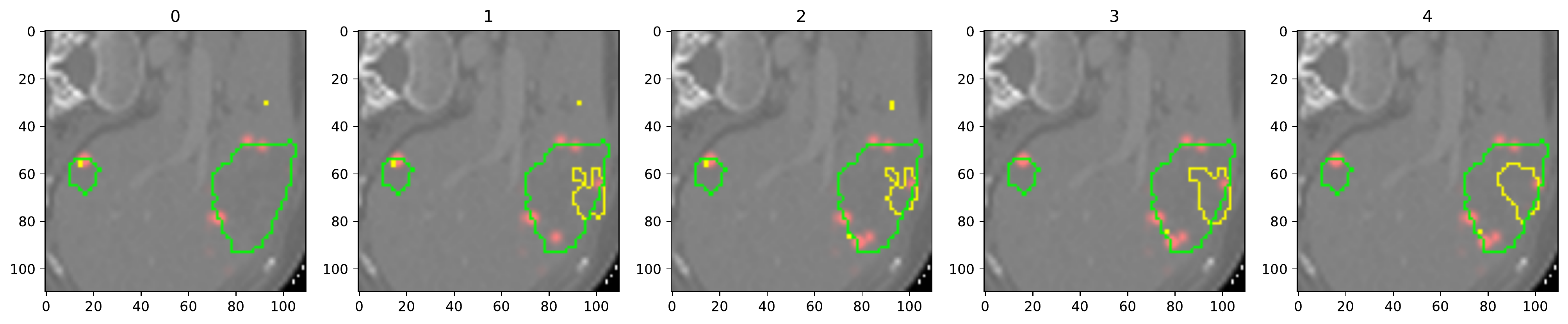}
}
\subfigure[InterCNN]{
\includegraphics[width=0.8\textwidth]{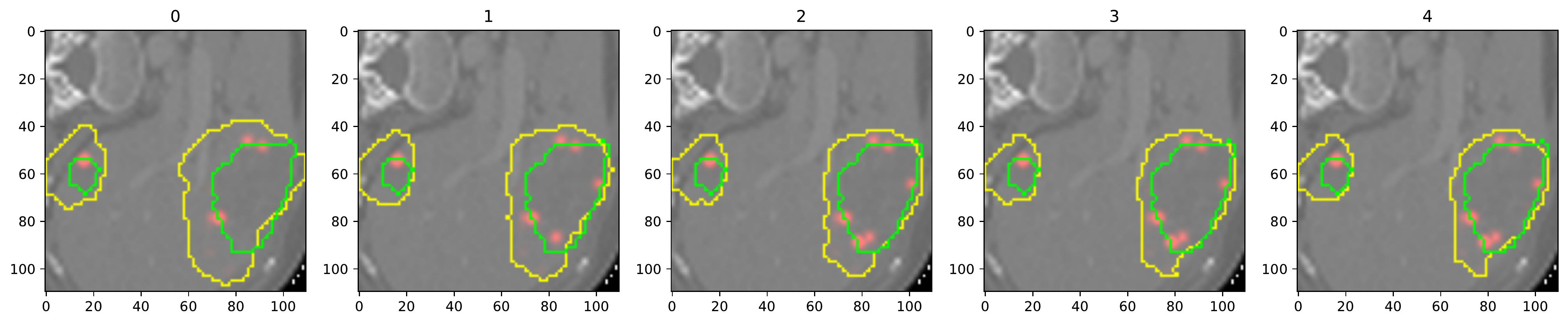}}
\subfigure[IteR-MRL]{
\includegraphics[width=0.8\textwidth]{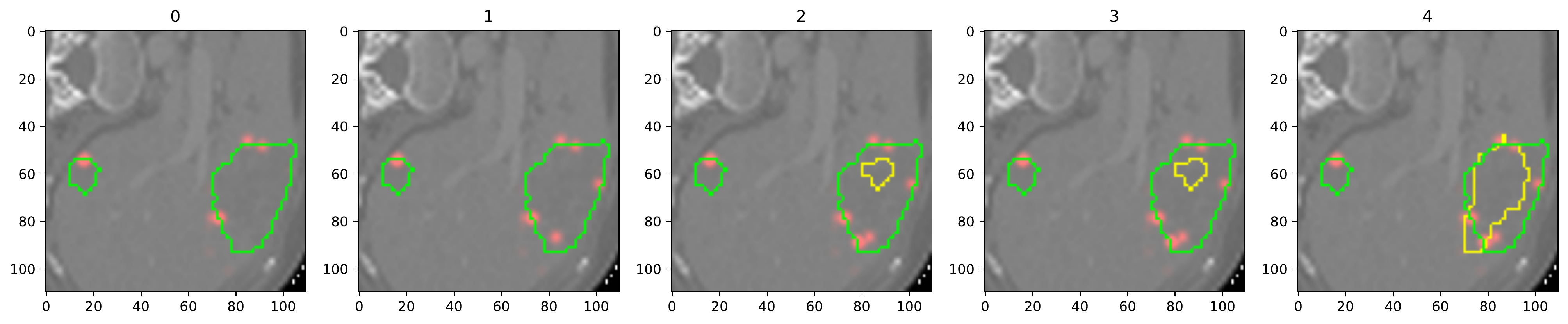}}
\subfigure[MECCA]{
\includegraphics[width=0.8\textwidth]{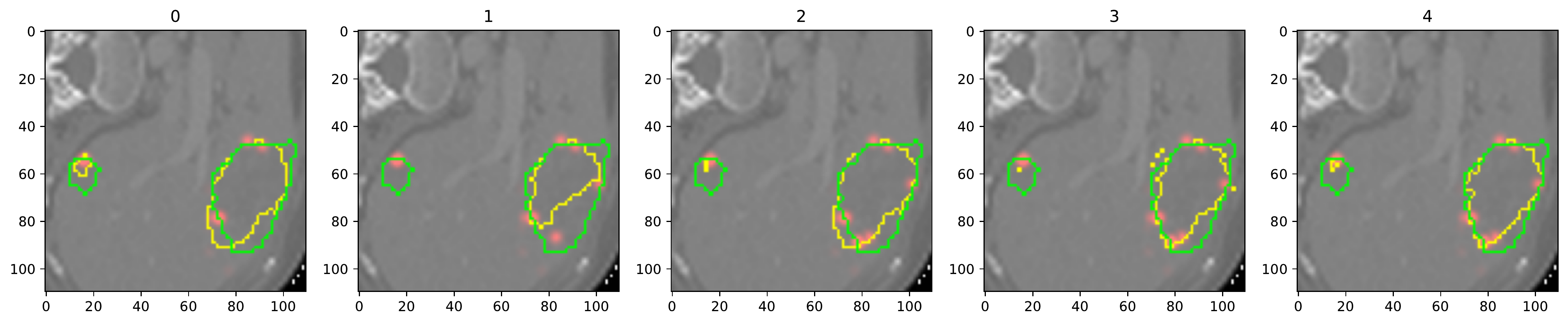}}
\caption{The results of different methods response to the same user interactions according to the same initial segmentation on 7/10 testing instance, 10/30 channel, Liver 2 of MSD dataset.}
\label{fig:sint3}
\end{figure}

Besides, we select the best and worst-performing samples from the test set for more analysis.
The results are shown in the Table~\ref{tab:difficults-vs-interacts}.
The improvement of easy samples through interaction is not considerable, but the improvement of hard samples through interaction can reach twice as much. 
However, significantly more interactions are required for hard samples to achieve the same segmentation accuracy of easy samples.
Moreover, it is very difficult and time-consuming to accurately mark edge points in a real scene, and is less practical to ask user to click the accurate edge points.
To verify MECCA's tolerance for inaccurate edge points, we conducted a robustness study under the same settings as Table~\ref{res1}.
Specifically, during each interaction in the training phase, random noises are added to simulated edge points. 
The noise range is plus or minus $2$ voxels in the three directions of $x$, $y$, and $z$. 
In this way, the edge points used by the algorithm will be randomly selected from $64\;(4\times 4\times 4)$ voxels within the real edge point neighborhood. 
This disturbance radius can cover the edge ambiguity area in most cases. 
In the testing phase, we also adopted the same disturbance operation, and the final results are shown in Table~\ref{tab:disturbed}.

From the table, it can be seen that the perturbed edge points will make the DICE value of the algorithm drop and have a more considerable variance.
However, the maximum value of DICE exceeds the accurate edge points. 
The perturbed edge points can even exceed the performance of accurate edge points in the ASSD value. 
We believe that there may be two reasons for the above phenomenon.
First, MECCA uses adaptive confidence calibration to improve the information-misunderstanding of the iterative algorithm, but it also makes the algorithm more ``conservative''. 
The perturbed edge point information may exceed the actual object boundary, which can make the segmentation effect of our method better or more ``radical'';
Second, perturbing the interactive information during the training phase can enable the reinforcement learning algorithm to explore the environment (medical images). 
Existing SOTA reinforcement learning algorithms generally impose entropy constraints on the policy, enhance the randomness of the policy, and encourage exploration.
Moreover, our perturbation of interactive information will indirectly affect the policy of the algorithm.
In a word, MECCA has good robustness to inaccurate edge points.

Finally, we show how different methods respond to the same user interactions according to the same initial segmentation, especially for hard cases.
We selected some images with poor performance in baselines or MECCA as the research objects.
Specifically, the generation mechanism of the same user interactions is described as following.
Total $45$ hint points are randomly selected from the intersection area of the boundary of the foreground object and the error region of the initialization segmentation.
Then, these $45$ points are allocated to the $5$ interactive steps of all methods according to the combination of $25$, $5$, $5$, $5$, and $5$.
The number of hint points used in each interactive step is the same as that of all methods during training and testing.
The results are shown in Figure~\ref{fig:sint1}-\ref{fig:sint3}.
It can be seen from the results that MECCA can use the hint points information stably in all cases.
The \textit{interactive misunderstanding} phenomenon arises in other methods.
These method either ignore the expert's correction information, or even be adversely affected by correction information as in figures.

\subsection{Comparison of Different Weighted Rewards}
\begin{figure}[ht!]
\centering
\includegraphics[width=0.8\textwidth]{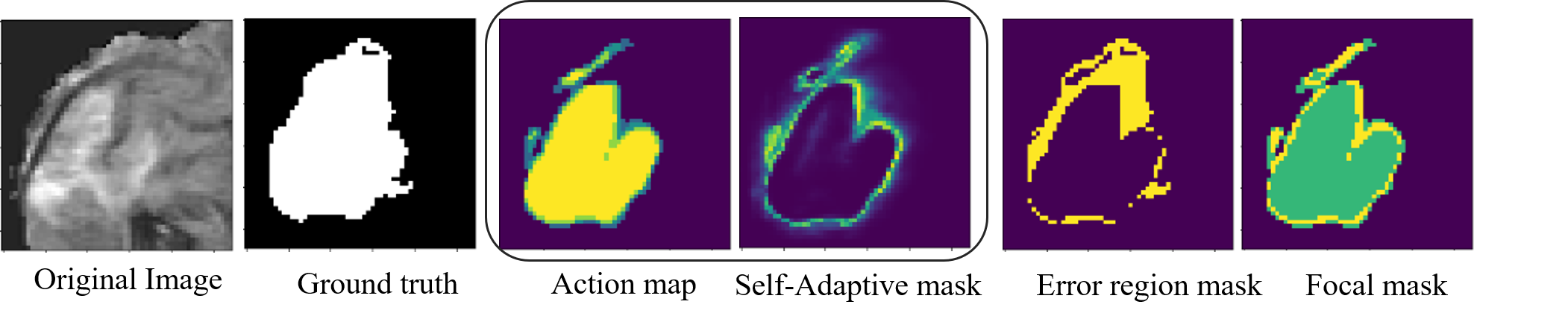}
\caption{Visualization of the self-adaptive mask, the Error region mask and the Focal mask. All masks are obtained at the first interaction step.
The brighter the color (closer to yellow), the larger the positive value; on the contrary, the darker the color (closer to black), the smaller the negative value.}
\label{maskview}
\end{figure}

\begin{figure*}[ht!]
\centering
\subfigure[]{
\label{segproc}
\includegraphics[width=0.8\textwidth]{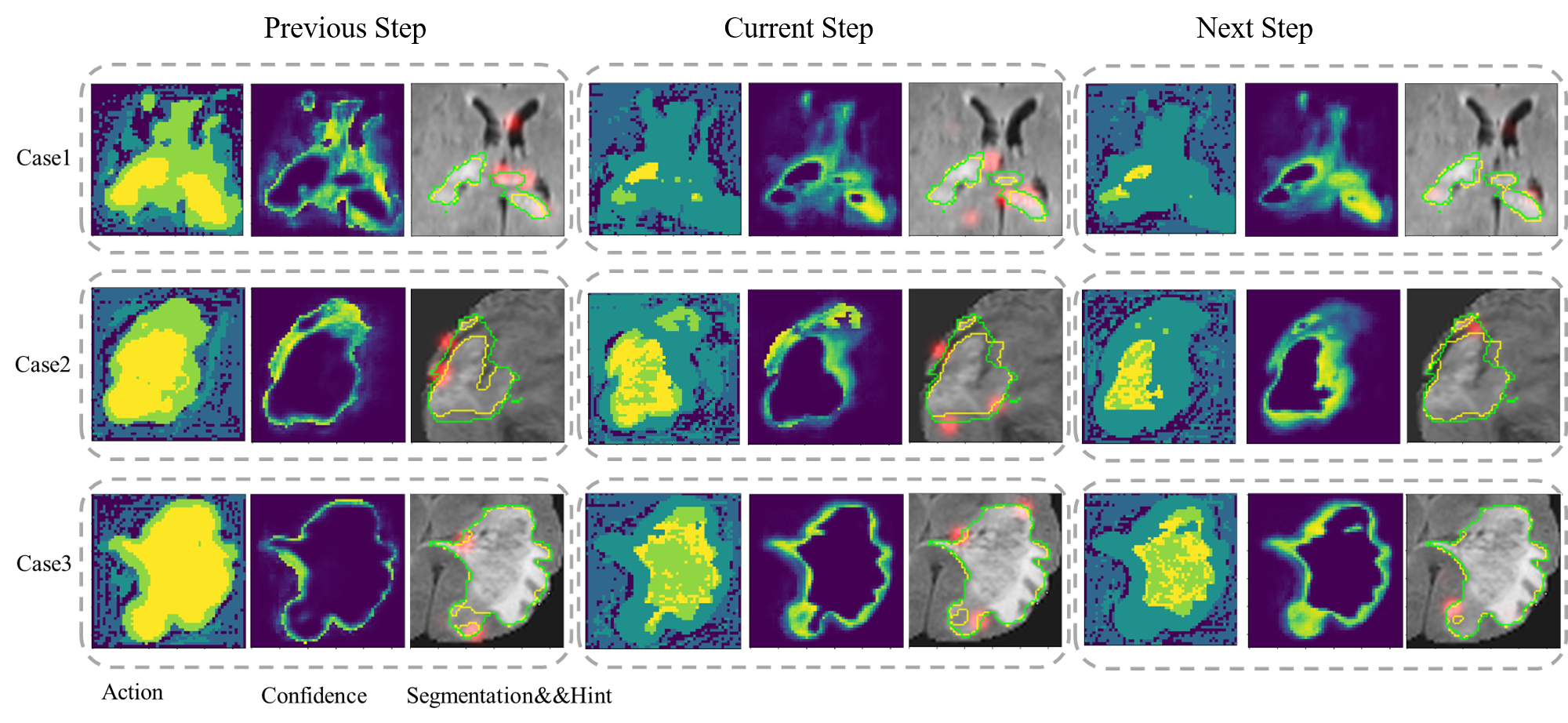}}
\subfigure[]{
\label{weightview}
\includegraphics[width=0.5\textwidth]{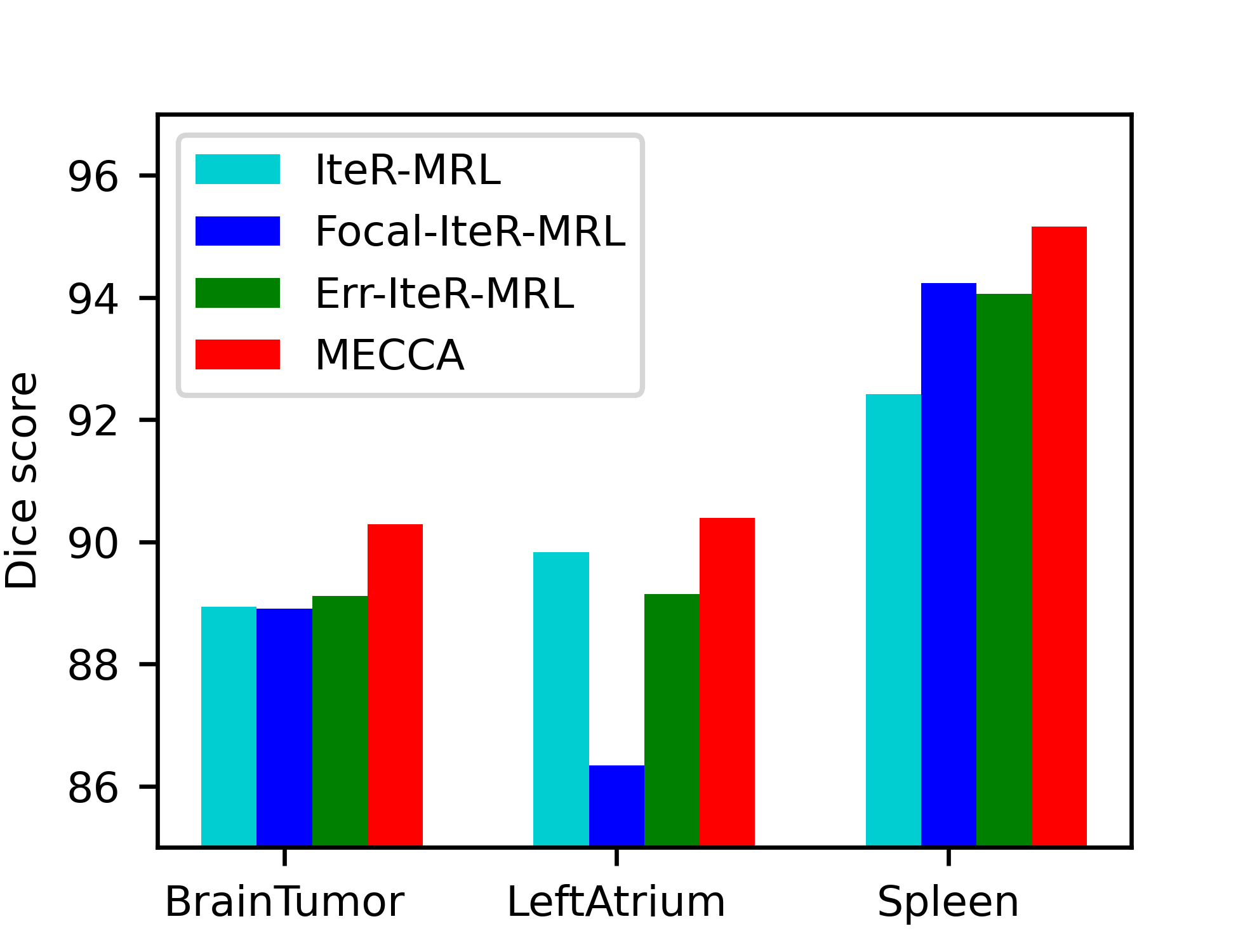}}
\caption{\textbf{(a)} The segmentation process of MECCA. At each step, the first column is the action map, the second column is the confidence map, and the third column is the segmentation result with the user's hint information.\textbf{(b)} Average Dice scores of methods with different weighting rewards. The reward function of IteR-MRL is not weighted; the reward function of Focal-IteR-MRL is weighted via (\ref{focalreward}), the reward function of Err-IteR-MRL is weighted in segmentation error regions, and the the reward function of MECCA is weighted through action confidence.
The brighter the color (closer to yellow), the larger the positive value; on the contrary, the darker the color (closer to black), the smaller the negative value.}
\end{figure*}

One main contribution of our work is that the self-adaptive reward mechanism is further proposed based on the confidence map. 
This mechanism makes different actions have different feedback levels so that the segmentation network can identify wrong actions as much as possible. 
To measure the impact of different reward weighting ways on the segmentation result, this study compares MECCA with the original IteR-MRL without weighting reward, the Focal-Reward IteR-MRL (Focal-IteR-MRL), and the one with weighted reward in segmentation error regions (Err-IteR-MRL).
The Focal-Reward IteR-MRL adapt the idea of Focal loss~\cite{lin2017focal}:
\begin{align} 
\label{focalreward}
FocalReward_{i}^{(t)} \ &=\ \mathcal{X}_{i}^{(t-1)} - \mathcal{X}_{i}^{(t)},
\end{align}
where
\begin{equation}
\mathcal{X}_{i}^{(t)}=\left\{\begin{array}{ll}
-(1-p_{i})y_{i}\log(p_{i}^{(t)}) & \text { if }  y_{i}=1, \\
-p_{i}(1-y_{i})\log(1-p_{i}^{(t)}) & \text { if } y_{i}=0.
\end{array}\right.
\end{equation}
The reward function of Err-IteR-MARL is a linear scaling of reward of IteR-MARL:
\begin{equation}
\text{ErrReward}_{i}^{(t)}= 
\begin{cases}  
(1 + \lambda_i) \text{gain}_{i}^{(t)}, & \text { if voxel i is misclassified,} \\ 
\text{gain}_{i}^{(t)}, & \text { otherwise, } \end{cases}
\end{equation}
where $\lambda_i$ is a positive real number hyperparameter.
The performance of different weighting rewards is shown in Figure~\ref{weightview}. 
Action-based reward weighting methods are more suitable for interactive segmentation since the improvement of our proposed method is more notable than other methods. 
Also, this study visualizes different masks that weigh the primary reward in Figure~\ref{maskview}. 
As we can see, the focal mask pays more attention to regions with low prediction probabilities, and the errors region mask contains all regions segmented incorrectly in the previous result. 
It is not enough for the focal mask to obtain more structured information by only considering the probability of prediction. 
As such, the performance of Focal-IteR-MRL is erratic, even the poorest on the left atrium dataset. 
The performance of Err-IteR-MRL is more stable than other methods, but the improvement is not notable. 
It is because that the weighting way of Err-IteR-MRL is based on the segmentation result while the refinement process of the interactive image segmentation is based on actions.

\begin{figure}[ht!]
\centering
\includegraphics[width=0.8\textwidth]{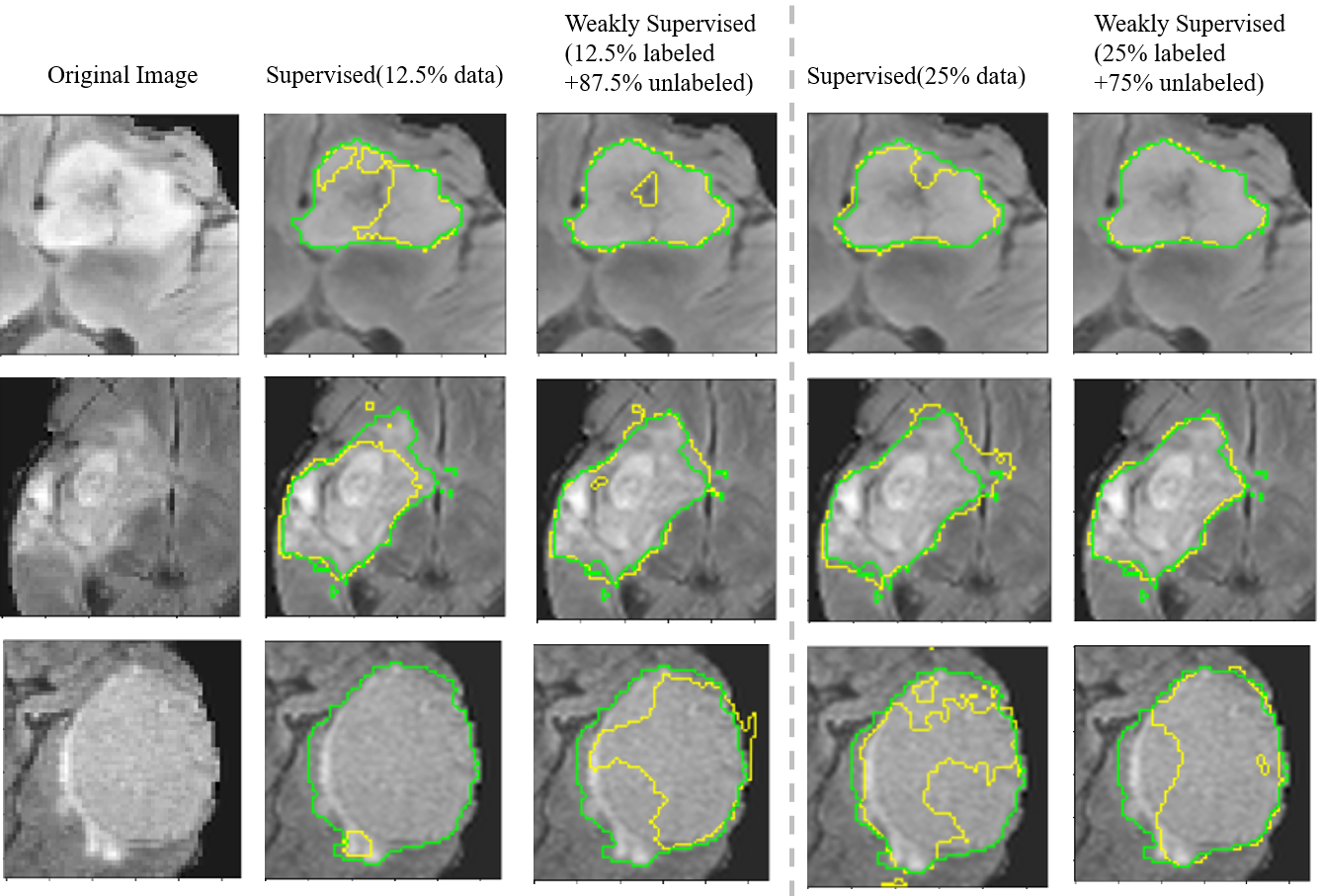}
\caption{Qualitative segmentation results of MECCA for BraTS2015 validation set. The first column shows the cropped original images. The second to fifth columns respectively show: the result of supervised learning only using $12.5$\% of labeled data, the results of weakly-supervised learning using $12.5$\% labeled and other unlabelled images, the result of supervised learning only using $25$\% of labeled data, the results of weakly-supervised learning using $25$\% labeled and other unlabelled images. Both supervised learning and weakly supervised learning is based on our proposed method.}
\label{weakview}
\end{figure}


\subsection{Weakly-supervised Interactive Segmentation}


\begin{table}[htbp]
\caption{Quantitative comparison between MECCA and other methods on BraTS2015 Dataset of different sizes. Significant improvement
(p-value $< 0.05$) is marked in bold.}
\label{weaktable}
\centering
 \begin{tabular}{p{30mm}|ccccccr}
\toprule[1pt]
Data amount & 1/8 & 1/4 & 1/3 & 1/2 & 1/1 & $\triangle(1/8, 1/1)$ \\
\midrule
P-Net     &75.86  &80.83  &80.88 &83.02 &84.00 & $10.73\%$\\
DeepIGeos &85.50  &85.90  &87.70 &87.60 &88.32 & $3.30\%$\\
IteR-MRL  &84.36  &86.54  &87.38 &88.66 &88.94  & $5.07\%$\\
UA-MT  & 83.08 & 84.47 & 84.62 & 84.39 & 84.66 & \textbf{1.90\%}\\
MECCA  &\textbf{87.14}  &\textbf{88.23}  &\textbf{88.31} &\textbf{89.17} &\textbf{90.29} & $3.60\%$ \\
\bottomrule
\end{tabular}
\end{table}

As mentioned in Section~\ref{unlabeled}, MECCA can reduce the dependence on voxel-level annotations of images by using the simulated label generated from the action confidence map. 
This study validates Algorithm~\ref{alg:1} on the BraTS2015 dataset by randomly selecting different proportions of samples as fully labeled data and using the rest of the training images as unlabeled data, which only provided hint information when interaction. 
This study compares MECCA against the P-Net (without hint), DeepIGeos, and IteR-MRL~\cite{liao2020iteratively}.
However, only MECCA uses unlabeled data, and the remaining three baselines only use a fixed proportion of labeled data.
This does not fully demonstrate the performance of MECCA on semi-supervised problems. 
For this reason, we additionally introduce a state-of-the-art semi-supervised method UA-MT~\cite{yu2018pu}.
\textcolor{black}{Table~\ref{weaktable} shows the results of different methods. 
MECCA achieves a Dice score of $87.14$\% with only $12.5$\% labeled data, and it performs better than the other methods with $25$\% labeled data.}

Since UA-MT is not an interactive segmentation algorithm, its absolute performance is worse than interactive segmentation baselines but better than the non-interactive method P-Net. 
In addition, it can be seen from the last column of results that UA-MT has the most negligible performance loss on different proportions of labeled data. 
After MECCA introduces the simulated label generation mechanism, the performance loss can also be controlled at a similar magnitude to UA-MT.
Interestingly, DeepIGeos can maintain a good performance loss without using a semi-supervised learning mechanism.
Combined with the performance loss of IteR-MRL, we can conclude that in the semi-supervised interactive segmentation task, the interactive misunderstanding phenomenon will exacerbate the performance loss caused by the missing data. 
DeepIGeos and MECCA alleviate the interactive misunderstanding phenomenon through hard constraints and self-adaptive confidence calibration, respectively, so both of them can achieve better results under semi-supervised settings.
However, it is worth mentioning that DeepIGeos does not consider multi-step interactions and the relationship between consecutive interactions. 
It cannot fully utilize long-term interactive information, so it has a large gap with MECCA in absolute performance.

\textcolor{black}{Figure~\ref{weakview} shows the qualitative segmentation results of our method trained with different data sizes. 
It shows that the model trained with $12.5$\% labeled data can only capture the main region of the tumor.
However, the model is unable to distinguish the infiltration areas of the tumor.
For instance, the boundaries of tumors in the figure are hard to distinguish as it is more similar to these healthy regions. 
In this case, the models trained with little data tend to ignore these boundary regions of the tumor, while the model trained with both labeled data and unlabeled data can get more smooth and accurate boundaries. 
This phenomenon is mainly due to the distribution of the training set is not consistent with the validation set. 
The main advantage of the model trained with both labeled and unlabeled data is that it can minimize the gap between training and validation data.}

\begin{table}[ht!]
\centering
\caption{DICE of the algorithm under different number of interactions and percentage of labeled data.}
\label{tab:labeled-vs-interacts}
\resizebox{0.6\textwidth}{!}{%
\begin{tabular}{|c|c|c|c|c|c|}
\hline
\rowcolor[HTML]{9B9B9B} 
\% Labeled Data & \multicolumn{5}{c|}{\cellcolor[HTML]{9B9B9B}Interaction Steps} \\ \hline
\rowcolor[HTML]{FFFFFF} 
& 3     & 4     & 5     & 6     & 7     \\ \hline
\rowcolor[HTML]{FFFFFF} 
25$\%$  & 84.03 & 86.18 & 88.23 & 87.32 & 88.24 \\ \hline
\rowcolor[HTML]{FFFFFF} 
33$\%$  & 85.32 & 87.60 & 88.31 & 87.99 & 88.53 \\ \hline
\rowcolor[HTML]{FFFFFF} 
50$\%$  & 85.39 & 87.07 & 89.17 & 88.93 & 89.13 \\ \hline
\rowcolor[HTML]{FFFFFF} 
100$\%$ & 89.93 & 90.40 & 90.29 & 89.85 & 90.79 \\ \hline
\end{tabular}%
}
\end{table}

We additionally test MECCA's performance under different interaction times and the percentage of labeled data, and the results are shown in Table~\ref{tab:labeled-vs-interacts}.
Based on the experimental results, we can obtain the following conclusions:
1) if there is enough labeled data, MECCA can achieve good results after a few interactions;
2) the number of interactions required by MECCA to achieve the same performance is roughly inversely proportional to the number of labeled data;
3) although MECCA requires more interactions (about $2$-$3$ times the number of interactions) when only part of the labeled data is available, it can approach the algorithm's performance trained with all labeled data in the end.

\subsection{Ablation Study}

{\begin{table}[htbp]
\caption{Ablation study of our proposed algorithm on the BraTS2015 dataset with $25\%$ labeled data and $75\%$ unlabeled data . Significant improvement (p-value < $0.05$) is marked in bold.} 
\label{ablation}
\centering
\begin{tabular}{p{18mm}p{20mm}p{20mm}|c}
\toprule[1pt]
Self-adaptive reward & Interaction guide & Simulated labels generation & Dice(\%) \\
\midrule
 &  &  &86.54  \\
\checkmark &  &  &87.21  \\
 &\checkmark  &  &86.56  \\
 &  &\checkmark  &88.01  \\
\checkmark &\checkmark  &  &87.24  \\
\checkmark &\checkmark  &\checkmark  &\textbf{88.23}  \\
\bottomrule
\end{tabular}
\end{table}}

{\begin{table}[htbp]
\caption{Ablation study of our proposed algorithm on the Liver 2 dataset of MSD with $25\%$ labeled data and $75\%$ unlabeled data. Significant improvement (p-value $<0.05$) is marked in bold.} 
\label{ablation-add}
\centering
\begin{tabular}{p{18mm}p{20mm}p{20mm}|c}
\toprule[1pt]
Self-adaptive reward & Interaction guide & Simulated labels generation & Dice(\%) \\
\midrule
 &  &  & 66.43 \\
\checkmark &  &  & 67.45 \\
 &\checkmark  &  & 66.56 \\
 &  &\checkmark  &  67.93 \\
\checkmark &\checkmark  &  & 67.59 \\
\checkmark &\checkmark  &\checkmark  &\textbf{69.99}  \\
\bottomrule
\end{tabular}
\end{table}}

Table \ref{ablation} and \ref{ablation-add} shows the impact of different mechanisms to the MECCA. 
It is clear that the stimulative labels generation mechanism and self-adaptive reward mechanism significantly improve the algorithm's performance. 
The stimulative labels generation mechanism can achieve a $1.47\%$ and $3.56\%$ gain over the initial algorithm without any mechanisms for the Brats2015 and Liver 2 dataset, respectively. 
The main reason for this significant improvement is the utilization of these unlabeled data.

\begin{table}[htbp]
\caption{Quantitative comparison of interactive segmentation methods in computational and interaction time.}
\label{Comtime}
\centering
 \begin{tabular}{p{25mm}|p{55mm}p{50mm}c}
\toprule[1pt]
Methods & Total inference time per iteration & Interaction time per iteration   \\
\midrule
DeepIGeos  &\textbf{510ms}  &226ms    \\
InterCNN   &532ms  &237ms    \\
IteR-MRL   &869ms  &235ms    \\
BS-IRIS    &990ms  &250ms    \\
MECCA   &631ms  &\textbf{123ms}    \\
\bottomrule
\end{tabular}
\end{table}

\begin{table}[htb!]
\centering
\caption{Quantitative comparison of interactive segmentation methods in interaction time on realistic segmentation platform operated by experts. Demo video of MECCA is available at \url{https://bit.ly/mecca-demo-video}.}
\label{tab:real-time}
\resizebox{0.8\textwidth}{!}{%
\begin{tabular}{|c|c|c|c|c|c|}
\hline
                    & \multicolumn{5}{c|}{Interaction Time (\textit{Second}, Real World)}                \\ \hline
Interaction Step(s) & 1 (25 points) & 2 (5 points) & 3 (5 points) & 4 (5 points) & 5 (5 points) \\ \hline
DeepIGeos & 70 & 15 & 15 & 8  & 8  \\ \hline
InterCNN  & 76 & 19 & 18 & 8  & 7  \\ \hline
IteR-MRL  & 75 & 17 & 17 & 8  & 8  \\ \hline
BS-IRIS   & 80 & 20 & 20 & 11 & 10 \\ \hline
MECCA     & \textbf{41} & \textbf{12} & \textbf{11} & \textbf{5}  & \textbf{5}  \\ \hline
\end{tabular}%
}
\end{table}

Table~\ref{Comtime} shows the comparison of interactive segmentation methods in computational and interaction time. 
The result shows that the MECCA only takes half the time of other methods when performing an interaction. 
The main contribution to the reduction of interaction time is the interaction guide mechanism mentioned in Section~\ref{inter guid}.
Furthermore, to explore the efficiency of MECCA in authentic tasks, we designed an interactive software platform. 
We asked oncologists to use it for natural interactive segmentation, and the results are in Table~\ref{tab:real-time}.
It can be seen from the results that the MECCA algorithm also takes only about half of the time of the baselines due to the interaction guidance.
In addition, as the segmentation results become more and more accurate, the interaction time required gradually decreases.

\section{Conclusion}
\label{conclusion}
This paper presents a novel action-based confidence learning method for interactive 3D image segmentation. 
Specifically, this paper proposes to learn the confidence of actions that continuously refine the segmentation result during the interaction process so that hint information can be used more effectively.
Based on this, the self-adaptive reward is proposed for the segmentation module, which can prevent the interactive misunderstanding phenomenon during the interaction and help reduce the time cost of interaction by providing users with the advice regions to interact next. 
Besides, the confidence map can also replace the ground truth to generate feedback for the unlabeled samples for the segmentation module. 
These samples without voxel-level annotations can also be used to train our model. 
By integrating these components, MECCA is demonstrated to medical segmentation efficiently via less voxel-level annotated samples.


\bibliography{main}
\bibliographystyle{main}

\end{document}